\documentclass{article}

\usepackage[main, final]{neurips_2026}

\usepackage[utf8]{inputenc} 
\usepackage[T1]{fontenc}    
\usepackage{hyperref}       
\usepackage{amsmath} 
\usepackage{url}            
\usepackage{booktabs}       
\usepackage{amsfonts}       
\usepackage{nicefrac}       
\usepackage{microtype}      
\usepackage{xcolor}         

\newcommand{\method}{\textit{4DThinker}\xspace}

\makeatletter
\renewcommand{\@noticestring}{Preprint.\\ The work was conducted during the internship of Zhangquan Chen at Meituan.}
\makeatother

\usepackage{graphicx}
\usepackage{float}
\usepackage{colortbl}
\usepackage{xspace}
\usepackage{enumitem}
\usepackage{multirow} 
\usepackage{booktabs, xcolor}
\usepackage{caption}
\title{\method: Thinking with 4D Imagery for \\ Dynamic Spatial Understanding}

\author{%
  Zhangquan Chen$^{1,2}$ \quad
  Manyuan Zhang$^{2,3\dagger}$ \quad
  Xinlei Yu$^{4}$ \quad
  Xiang An$^{5}$ \quad
  Bo Li$^{5}$ \quad
  Xin Xie$^{3}$ \\
  \textbf{ZiDong Wang$^{2,3}$ \quad
  Mingze Sun$^{1}$ \quad
  Shuang Chen$^{6}$ \quad
  Hongyu Li$^{2,3}$ \quad
  Xiaobin Hu$^{4}$ \quad
  Ruqi Huang$^{1\dagger}$} \\[0.5em]
  $^{1}$Tsinghua University, SIGS \quad
  $^{2}$Meituan \quad
  $^{3}$The Chinese University of Hong Kong \\
  $^{4}$National University of Singapore \quad
  $^{5}$LMMs-Lab \quad
  $^{6}$University of California, Los Angeles \quad \\[0.3em]
  $^{\dagger}$Corresponding authors \\
}

\begin{document}
\maketitle
\begin{abstract}
Dynamic spatial reasoning from monocular video is essential for bridging visual intelligence and the physical world, yet remains challenging for vision-language models (VLMs). Prior approaches either verbalize spatial-temporal reasoning entirely as text, which is inherently verbose and imprecise for complex dynamics, or rely on external geometric modules that increase inference complexity without fostering intrinsic model capability. In this paper, we present \method, the first framework that enables VLMs to ``think with 4D'' through \emph{dynamic latent mental imagery}, i.e., internally simulating how scenes evolve within the continuous hidden space. Specifically, we first introduce a \emph{scalable, annotation-free} data generation pipeline that synthesizes 4D reasoning data from raw videos. We then propose Dynamic-Imagery Fine-Tuning (\textbf{DIFT}), which jointly supervises textual tokens and 4D latents to ground the model in dynamic visual semantics. Building on this, 4D Reinforcement Learning (\textbf{4DRL}) further tackles complex reasoning tasks via outcome-based rewards, restricting policy gradients to text tokens to ensure stable optimization. Extensive experiments across multiple dynamic spatial reasoning benchmarks demonstrate that \method consistently outperforms strong baselines and offers a new perspective toward 4D reasoning in VLMs. Our code is available at \url{https://github.com/zhangquanchen/4DThinker}.
\end{abstract}

\section{Introduction}
\label{sec:intro}

The physical world is inherently dynamic. For an intelligent agent to truly understand the environment, it must go beyond static perception and reason about \emph{how things change in 3D space over time}. 
Dynamic spatial reasoning, the ability to decompose and interpret the interplay of camera ego-motion and object motion from monocular video, is therefore a cornerstone of real-world visual intelligence, with direct implications for autonomous driving, and robotics~\cite{dsibench,liao2026spamem}.

Despite rapid advances in vision-language models (VLMs), recent benchmarks expose that even the strongest models fail at basic dynamic reasoning~\cite{vlm4d,dsibench}. Existing efforts to close this gap broadly follow two directions. One constructs 4D post-training data that verbalizes spatial-temporal reasoning entirely as text~\cite{llava4D,zhu2026egoreasoner,thinkdynamic}, yet natural language is \emph{inherently verbose and struggles to precisely convey complex dynamics}~\cite{latentsurvey}. The other augments the model with external modules, e.g., injecting geometric priors via 3D foundation models~\cite{dsrsuite} or appending mask decoders for spatial grounding~\cite{videoloom}, but \emph{at the cost of increased inference complexity and non-intrinsic model capability}. 
A promising alternative is  \emph{latent reasoning}~\cite{latentsurvey}, which encodes reasoning cues in continuous hidden space rather than explicit tokens. 
However, existing latent methods are limited to static scenes and depend on annotated reference images or distilled foundation models for supervision, \emph{hindering their scalability to the dynamic, annotation-scarce video domain.}

These limitations motivate three core desiderata:
\textbf{(D1)~Imagery-Dynamic}: extend latent visual reasoning beyond static scenes to \emph{capture 4D spatial-temporal evolution};
\textbf{(D2)~Model-Intrinsic}: embed reasoning capabilities directly \emph{within the model}, obviating the need for external geometric modules;
\textbf{(D3)~Data-Scalable}: scale the training paradigm \emph{without relying on manual annotations}.

We take inspiration from how humans naturally reason about motion. When observing a dynamic scene, we naturally parse motion by mentally simulating salient landmarks, i.e., anchoring on static cues to infer ego-motion, and tracking trajectories to understand object dynamics. \method operationalizes this insight by highlighting salient landmarks via mask overlays and \emph{treating these highlighted frames as  ``imagery'' that the model learns to simulate within its latent space.}

Specifically, \method introduces a ``think with 4D'' framework with three components. \emph{First}, we design a \emph{scalable, annotation-free data generation pipeline} that synthesizes 4D reasoning data directly from raw videos~(\textbf{D3}). The pipeline decomposes dynamic understanding along camera-motion and object-motion axes, generating motion-centric QA pairs with chain-of-thought that interleaves textual analysis and dynamic mental imagery. \emph{Second}, we propose \emph{Dynamic-Imagery Fine-Tuning} (DIFT), a supervised training that grounds the model's intrinsic 4D latents in dynamic visual semantics~(\textbf{D1, D2}). DIFT jointly optimizes a cross-entropy loss on text tokens and a cosine-similarity loss on latent positions, teaching the model to \emph{internally simulate} dynamics over time. \emph{Third}, we introduce \emph{4D Reinforcement Learning} (4DRL), a modified GRPO training that addresses challenging motions via outcome-based rewards~(\textbf{D1}). 
The policy gradient is restricted to text tokens only, excluding latent positions where continuous hidden-state propagation is misaligned with discrete log-probabilities.

Our contributions can be summarized as follows.
\begin{itemize}
\item We propose \method, the first ``think with 4D'' framework that equips VLMs with the capacity to \emph{mentally simulate 4D dynamics}, enabling intrinsic reasoning about camera and object motion without any external geometric modules.
\item We introduce a \emph{scalable, annotation-free} pipeline that synthesizes 4D reasoning data from raw videos, featuring Chain-of-Thought (CoT) interleaved with dynamic mental imagery.
\item We design a two-stage training recipe: \emph{DIFT} jointly supervises text and dynamic imagery for reasoning warm-up, while \emph{4DRL} selectively optimizes text tokens only, further refining compound-motion reasoning through outcome-based rewards.
\item Extensive experiments across multiple benchmarks demonstrate that \method consistently outperforms strong baselines, validating its effectiveness in dynamic spatial reasoning.
\end{itemize}

\section{Related Work}
\label{sec:related}

\subsection{Latent Reasoning}
\label{sec:related_latent}


Chain-of-thought (CoT) prompting~\cite{wei2022chain,chen2026omnivideo,jiang2026beyond,xu2025learning,lan2025contextual} has proven effective at eliciting multi-step reasoning from large language models (LLMs) by verbalizing intermediate steps. 
However, verbal reasoning is inherently redundant in tokens and \emph{struggles to precisely convey complex spatial-temporal cues (e.g., 3D layouts, dynamic trajectories)~\cite{latentsurvey, chen2025sifthinker}.}
This motivates \emph{latent reasoning}, which shifts part of the reasoning from the explicit token space into the model's continuous hidden space. 

Early explorations introduced dedicated tokens to structure the latent computation. 
Pause-pretraining~\cite{goyal2024think} inserts learnable \texttt{<pause>} tokens that grant extra computation steps before committing to output.
Implicit CoT~\cite{deng2024explicit} distills explicit CoT traces into implicit hidden-state trajectories, internalizing reasoning without explicit token generation.
COCONUT~\cite{coconut} goes further by replacing CoT tokens entirely with continuous latent embeddings, showing that multi-hop reasoning paths can be effectively encoded on a continuous manifold.
Moving beyond text-only models, recent work has explored latent reasoning in multimodal settings. Mirage~\cite{mirage} introduces \emph{machine mental imagery}, interleaving compact latent visual tokens with text by recasting hidden states as visual embeddings. 
LVR~\cite{lvr} similarly employs latent visual tokens but performs intrinsic iterative refinement without auxiliary image supervision. 
Most recently, 3DThinker~\cite{3dthinker} extends this paradigm to 3D, aligning latent tokens with a 3D foundation model to enable geometric imagination during spatial reasoning.


Despite these advances, existing methods remain confined to pure text, 2D images, or static scenes. \method is \emph{the first to extend latent visual tokens to spatial-temporal dynamics}, enabling the model to internally simulate object trajectories, camera motion, and their interplay in video.

\subsection{Visual-Spatial Understanding}
\label{sec:related_spatial}
Spatial understanding has received increasing attention as a core capability for models interacting with the 3D world~\cite{chen2024spatialvlm,cai2024spatialbot,chan2026adagar,li2025slam,li2023hong,hou2025federated,yin2026mllm,li2025llava}. On the benchmark side, a series of works~\cite{cvbench,vsibench,ma20243dsrbench} systematically probe spatial competencies such as distance estimation and relative positioning. On the modeling side, one line of methods augment inputs with explicit geometric signals, i.e., depth maps~\cite{rela:liu2025spatialcot}, or point clouds~\cite{vlm3r}, to supply 3D priors directly. Another enhances intrinsic reasoning without external geometry, e.g., MindCube~\cite{mindcube} constructs textual cognitive maps while 3DThinker~\cite{3dthinker} generates latent 3D tokens for geometric imagination. Despite notable progress, these efforts remain largely confined to \emph{static} scenes, with dynamic spatial reasoning in video still underexplored.
Extending spatial reasoning to dynamic scenes from monocular video, where both the camera and objects may move, poses a harder and more practically relevant challenge. 
VLM4D~\cite{vlm4d} first highlights this gap with a benchmark showing that even strong VLMs fail at basic motion direction reasoning.
DSI-Bench~\cite{dsibench} further decouples camera and object motion, revealing that VLMs systematically conflate the two.
More recently, DSR Suite~\cite{dsrsuite} provides a large-scale dataset alongside a Geometry Selection Module (GSM) that injects geometric priors for dynamic spatial reasoning. VideoLoom~\cite{videoloom} introduces SlowFast token designs that decouple temporal context from spatial detail for joint spatial-temporal understanding.

However, existing methods for dynamic spatial understanding all rely on external geometric modules that \emph{increase inference complexity}.
\method enables the model to \emph{internally simulate} through latent visual tokens, \emph{without additional modules or priors.} That is, \method develops an intrinsic capacity for 4D reasoning, \emph{arriving at answers through ``mental imagery'' of the dynamic scene.}

\begin{figure*}[]
    \centering
    \includegraphics[width=\textwidth]{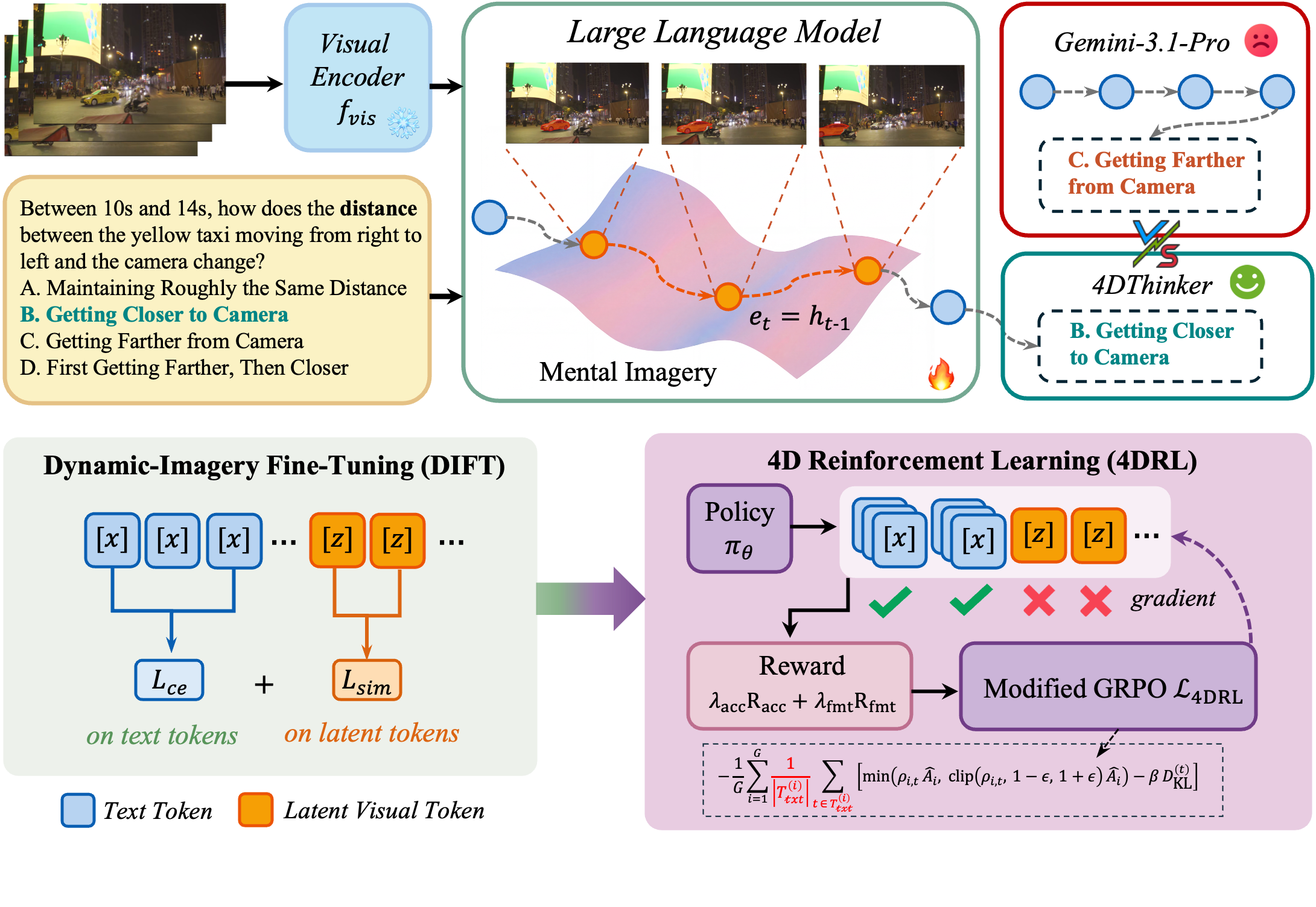}
    \vspace{-3.5em}
    \caption{\textbf{Overview of \method.} \textbf{Top:} Inference architecture. The model interleaves text reasoning with \emph{latent visual tokens} as ``mental imagery'' on a continuous manifold, enabling correct dynamic reasoning where purely textual CoT (e.g., Gemini-3.1-Pro) fails. \textbf{Bottom:} Two-stage training pipeline built on the data from Fig.~\ref{fig:data_gen}. DIFT (left) jointly supervises text tokens via $\mathcal{L}_{\text{ce}}$ and latent tokens via $\mathcal{L}_{\text{sim}}$; 4DRL (right) then applies modified GRPO with gradients on text tokens only, excluding latent positions to avoid noise from continuous-discrete mismatch.}\label{fig:pipeline}
    \vspace{-0.8em}
\end{figure*}
\section{Methodology}
\label{sec:method}
Understanding dynamic scenes from monocular video requires reasoning about how objects and the camera move through 3D space over time. Inspired by the cognitive mechanism of mental imagery, we propose \method, a framework that enables VLMs to \emph{internally visualize spatial-temporal dynamics during reasoning via latent visual tokens, without relying on any external geometric modules.} As illustrated in Fig.~\ref{fig:pipeline}, \method consists of three key components: (1) a \emph{scalable, annotation-free} data generation pipeline that synthesizes 4D reasoning data from raw videos (Sec.~\ref{sec:data}); (2) \emph{Dynamic-Imagery Fine-Tuning} (DIFT), which grounds 4D latents in dynamic visual semantics through joint supervision (Sec.~\ref{sec:training}); and (3) \emph{4D Reinforcement Learning} (4DRL), which further refines reasoning on complex compound motions via outcome-based rewards (Sec.~\ref{sec:training}).

\subsection{Scalable 4D Data Generation}
\label{sec:data}
\begin{figure*}[]
    \centering
    \includegraphics[width=\textwidth]{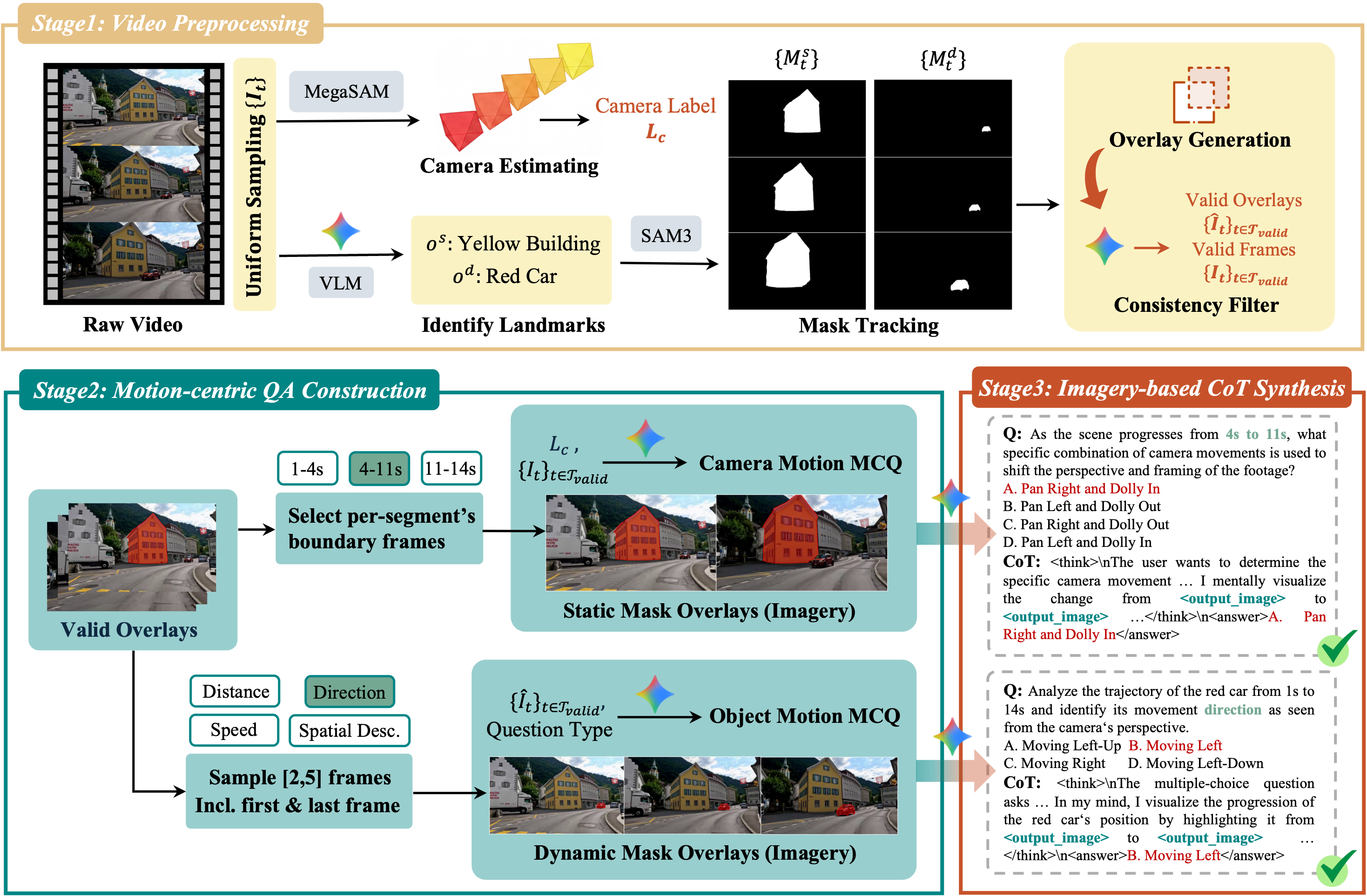}
    \vspace{-1.4em}
    \caption{Overview of our \textbf{scalable, annotation-free} 4D data generation pipeline in three stages. \textbf{(1) Video preprocessing:} raw videos are processed via MegaSaM and SAM3 to extract camera trajectories and consistent mask overlays for landmarks. \textbf{(2) Motion-centric QA construction:} the pipeline formulates MCQs and imagery for both camera and object motions, grounded by sampled boundary or interval overlays. \textbf{(3) Imagery-based CoT synthesis:} $M_\text{high}$ generates ``think with 4D'' data that interleaves text and dynamic mental imagery, culminating in the final training sample.}\label{fig:data_gen}
    \vspace{-1.4em}
\end{figure*}

Manual annotation for spatial-temporal understanding data is \emph{expensive} and inherently \emph{unscalable}. On the other hand, our method requires reformulating conventional QA data into CoT reasoning \emph{grounded in dynamic mental imagery}. 
To bridge this gap, we propose a \emph{scalable, annotation-free} pipeline to synthesize 4D reasoning data from raw videos. This pipeline sequentially executes video preprocessing, motion-centric QA construction, and imagery-based CoT synthesis as shown in Fig.~\ref{fig:data_gen}.

\paragraph{Video preprocessing.}
Our training corpus is built from SpatialVID~\cite{spatialvid}, a large-scale video collection. We use only its videos and geometric annotations estimated automatically by MegaSaM~\cite{megasam}, introducing \emph{no information that requires human annotation}.

Since dynamic reasoning relies on salient objects to gauge relative motion, our first step is to identify these landmarks and extract their masks. 
Given a video $\mathcal{V}$, we uniformly sample frames to obtain $\{I_t\}_{t=0}^{T-1}$, where $T$ is the video duration in seconds.
Based on predefined rules $\mathcal{R}$ (see Appendix~\ref{app:rules}), we query a high-level model $M_{\text{high}}$ (e.g., Gemini3-pro) to identify a representative \emph{static} object $o^s$ (e.g., the red building) and a \emph{dynamic} object $o^d$ (e.g., the person riding the blue bike) that persist throughout the video.
A promptable video segmentation model (SAM3~\cite{sam3}) then tracks each object across all frames, producing temporally consistent binary mask sequences $\{M^s_t\}$ and $\{M^d_t\}$. Leveraging these masks, we generate the \emph{mask overlays} to highlight the target object:
\begin{equation}
\label{eq:overlay}
\hat{I}_t = (1 - \alpha \cdot M_t) \odot I_t \;+\; \alpha \cdot M_t \odot \mathbf{c},
\end{equation}
where $\alpha \in (0,1]$ is the opacity, $\mathbf{c} \in \mathbb{R}^3$ is the highlight color, and $\odot$ denotes element-wise multiplication.
To prevent identity drift, we also apply a \emph{consistency filter}. Specifically, we prompt $M_{\text{high}}$ to cross-verify the entire set, retaining only temporally consistent overlays:
\begin{equation}
\label{eq:filter}
\mathcal{T}_{\text{valid}} = \big\{\, t \;\big|\; \Phi_{M_{\text{high}}}\!\big(\hat{I}_t,\;\{\hat{I}_{t'}\}_{t'\neq t}\big) = \texttt{True} \,\big\}.
\end{equation}
This yields a set of \emph{valid overlays} $\{\hat{I}_t\}_{t \in \mathcal{T}_{\text{valid}}}$ together with their corresponding \emph{valid frames} $\{I_t\}_{t \in \mathcal{T}_{\text{valid}}}$, which serve as the foundation for all subsequent data construction.


After preprocessing, we decouple dynamic understanding for monocular videos into \emph{camera motion} and \emph{object motion}, and structure our data generation along these two axes.

\paragraph{Camera motion data.}
From the camera trajectories produced by MegaSaM, SpatialVID derives per-segment camera motion labels $L_c$ covering 12 canonical movement types. We first partition the video timeline into temporally contiguous segments based on these labels, yielding a sequence of labeled intervals $\{[t_a^i,\,t_b^i]\}_{i=1}^{M}$. For a given segment $[t_a, t_b]$ within this sequence, we leverage its associated motion label in conjunction with the valid images $\{I_t\}_{t \in \mathcal{T}_{\text{valid}}}$ to prompt $M_\text{high}$, formulating the camera motion Multiple-Choice Question (MCQ), denoted as $(Q^s, A^s)$.

To establish the corresponding visual imagery, for each labeled segment $[t_a,\,t_b]$, we extract the \emph{static mask overlays} at the boundary frames from the valid set $\{\hat{I}^s_t\}_{t \in \mathcal{T}_{\text{valid}}}$ (Eq.~\eqref{eq:filter}), yielding $\hat{I}^s_{t_a}$ and $\hat{I}^s_{t_b}$. The key insight here is that for a static object ($\Delta\mathbf{p}_t^{\,\text{obj}} = \mathbf{0}$), its apparent displacement in the image plane is entirely attributable to camera motion:
\begin{equation}
\label{eq:static_disp}
\bar{\mathbf{p}}^{\,s}_{t_b} - \bar{\mathbf{p}}^{\,s}_{t_a} \;=\; \Delta\mathbf{p}^{\,\text{cam}}_{[t_a,\,t_b]},
\end{equation}
where $\bar{\mathbf{p}}^{\,s}_{t}$ denotes the centroid of $M^s_t$ in image coordinates. Consequently, these boundary overlays serve as explicit visual evidence of camera movement. 

Ultimately, the MCQ and imagery components are aggregated, culminating in the complete sample $s^s$: $\big(Q^s,\;A^s,\;\{I_t\}_{t \in \mathcal{T}_{\text{valid}}},\;\{\hat{I}^s_{t_a},\hat{I}^s_{t_b}\}\big)$. Additional details are provided in the Appendix~\ref{app:prompts}.




\paragraph{Object motion data.}

For the dynamic object $o^d$, we formulate candidate question types encompassing direction, distance, speed, and spatial descriptions grounded by bounding boxes (derived from masks). 
To deduce the ground-truth motion attributes, we prompt $M_{\text{high}}$ with the valid overlays of dynamic object $\{\hat{I}^d_t\}_{t \in \mathcal{T}_{\text{valid}}}$ (Eq.~\eqref{eq:filter}) to analyze the trajectory (see Appendix~\ref{app:prompts}), explicitly accounting for both in-plane displacements and apparent scale variations. By integrating these trajectory analyses with the predefined question types, we construct the object motion MCQ, denoted as $(Q^d, A^d)$. 

Concurrently, to establish the visual imagery, we generate the \emph{dynamic mask overlays} $\{\hat{I}^d_{t_i}\}_{i=1}^{N}$ by sampling $N \in [2, 5]$ frames from the valid overlays.
To capture the complete motion extent, this sampling process mandates the inclusion of the first and last frames of the object's active interval.
The final object motion sample $s^d$ is thus formulated: $\big(Q^d,\;A^d,\;\{I_t\}_{t \in \mathcal{T}_{\text{valid}}},\;\{\hat{I}^d_{t_i}\}_{i=1}^{N}\big)$.

\paragraph{Imagery-based CoT synthesis.}

We argue that \emph{discriminating complex motion inherently requires mentally visualizing the temporal dynamics of the attended object.} 
To emulate this, we synthesize structured CoT reasoning that interleaves textual analysis with \emph{dynamic mental imagery}. 
Given the previously formulated $s^s$ or $s^d$, $M_{\text{high}}$ is prompted to produce a ``think with 4D'' reasoning trace $r$.
Specifically, each CoT $r$ adheres to a structured format: \texttt{<think>...<imagery>..<imagery>...</think><answer>...</answer>}.
An automated validator subsequently verifies the placeholder count, chronological consistency, and answer isolation; non-compliant samples are either regenerated or discarded. Additional details are in the Appendix~\ref{app:cot}.

\paragraph{Training data composition.}
Executing the proposed pipeline yields ${\sim}$38K pairs tailored for supervised training. Each sample encapsulates CoT with mental imagery, formally defined as:
\begin{equation}
\label{eq:sample}
\mathcal{S} = \Big(Q,\;A,\;\{I_t\}_{t \in \mathcal{T}_{\text{valid}}},\;\{\hat{I}_{t_i}\},\;\; r\,\Big).
\end{equation}
While the supervised corpus warms up reasoning on single-category motions, we introduce an RL stage using DSR-Train (${\sim}$37K samples)~\cite{dsibench} to master complex, compound motions. Lacking explicit reasoning traces, this QA-only dataset compels the model to autonomously explore reasoning paths, guided solely by outcome-based rewards.

\subsection{Learning to Think with 4D}
\label{sec:training}
Building upon the dataset generated in Sec.~\ref{sec:data}, we now describe how \method learns to internalize dynamic imagery as part of its reasoning process.
That is, we represent mental imagery as \emph{latent visual tokens}, the compact continuous embeddings that reside within the hidden space of the language model. 
We first formalize this representation, then present a two-stage training framework: dynamic-imagery fine-tuning (DIFT), followed by 4D reinforcement learning (4DRL).

\paragraph{Latent visual token representation.}
Let $f_{\text{vis}}$ denote the visual encoder of the base VLM.
For an overlay image (i.e., imagery) $\hat{I}_{t_i}$ (Eq.~\eqref{eq:overlay}), the encoder produces a patch-level embedding sequence $\mathbf{E}_{t_i} = f_{\text{vis}}(\hat{I}_{t_i}) \in \mathbb{R}^{L \times D}$, where $L$ is the number of visual tokens and $D$ is the hidden dimension.
We compress this sequence into $K$ \emph{latent visual tokens} via partitioned mean pooling:
\begin{equation}
\label{eq:compress}
\mathbf{z}_{t_i}^{(k)} = \frac{1}{|\mathcal{P}_k|}\sum_{j \in \mathcal{P}_k} \mathbf{E}_{t_i}[j], \quad k = 1,\ldots,K,
\end{equation}
where $\{\mathcal{P}_k\}_{k=1}^{K}$ is an equal partition of $\{1,\ldots,L\}$.
Each \texttt{<imagery>} placeholder (i.e., special token) in the CoT $r$ is then replaced by a \emph{latent block}:
\begin{equation}
\label{eq:latent_block}
\texttt{<lat\_s>}\;\; \mathbf{z}^{(1)}_{t_i}\;\; \mathbf{z}^{(2)}_{t_i}\;\; \cdots\;\; \mathbf{z}^{(K)}_{t_i}\;\; \texttt{<lat\_e>},
\end{equation}
where \texttt{<lat\_s>} and \texttt{<lat\_e>} serve as learnable delimiter tokens. 
Consequently, the training sequence interleaves discrete text tokens with continuous latent blocks, enabling the model to \emph{reason through dynamic imagery} without leaving the autoregressive generation loop.

\paragraph{Dynamic-imagery fine-tuning (DIFT).}
Given the sample $\mathcal{S}$ (Eq.~\eqref{eq:sample}), we form the input by encoding video frames $\{I_t\}$ as visual tokens, appending the question $Q$, and substituting each imagery placeholder in $r$ with its corresponding latent block.
The visual encoder $f_{\text{vis}}$ is kept frozen throughout this stage, providing a stable target embedding space.
We optimize a dual-objective loss:
\begin{equation}
\label{eq:sft_loss}
\mathcal{L}_{\text{DIFT}} = \lambda_{\text{ce}}\,\mathcal{L}_{\text{ce}} \;+\; \lambda_{\text{sim}}\,\mathcal{L}_{\text{sim}}.
\end{equation}
The first term is the standard causal language modeling loss restricted to text token positions $\mathcal{T}_{\text{txt}}$:
\begin{equation}
\label{eq:ce_loss}
\mathcal{L}_{\text{ce}} = -\frac{1}{|\mathcal{T}_{\text{txt}}|}\sum_{t \in \mathcal{T}_{\text{txt}}} \log\, p_\theta\!\left(x_{t+1} \mid x_{\leq t}\right).
\end{equation}

The second term introduces a \emph{next-embedding prediction} objective at latent positions.
Let $\mathcal{T}_{\text{lat}}$ denote the set of all latent token positions and $\mathbf{h}_t$ the hidden state at position $t$.
Adhering to the autoregressive paradigm, $\mathbf{h}_{t-1}$ serves as the predictive representation for position $t$; we enforce its alignment with the ground-truth visual embedding $\mathbf{z}_t$ via cosine similarity:
\begin{equation}
\label{eq:sim_loss}
\mathcal{L}_{\text{sim}} = 1 - \frac{1}{|\mathcal{T}_{\text{lat}}|}\sum_{t \in \mathcal{T}_{\text{lat}}} \frac{\mathbf{h}_{t-1}^{\top}\,\mathbf{z}_t}{\|\mathbf{h}_{t-1}\|\;\|\mathbf{z}_t\|}.
\end{equation}
Essentially, this objective imparts 4D patterns through continuous supervision, i.e., the model learns to \emph{internally simulate the visual dynamics of attended objects at each imagery step.}

During inference, the DIFT formulation naturally gives rise to a \emph{recurrent} mental imagery mechanism that operates in a purely self-conditioned manner.
This is achieved by directly feeding the preceding hidden state as the input embedding whenever the current position falls within a latent block:
\begin{equation}
\label{eq:autoreg_vis}
\mathbf{e}_t = \begin{cases}
\operatorname{Embed}(x_t), & t \notin \mathcal{T}_{\text{lat}}, \\[4pt]
\mathbf{h}_{t-1}, & t \in \mathcal{T}_{\text{lat}},
\end{cases}
\end{equation}
where $\operatorname{Embed}(\cdot)$ denotes the standard discrete token embedding lookup. 
This establishes a recurrent loop: \emph{the model's own ``imagination'' at one imagery step feeds forward as context for subsequent reasoning, allowing it to mentally track how objects move in 3D space over time.}

\paragraph{4D reinforcement learning (4DRL).}
Although DIFT equips the model with the ability to reason via dynamic imagery, the supervised signal is limited to single-category motion, leaving its understanding of complex 4D scenes somewhat constrained. 
To overcome this limitation, we further apply a modified version of GRPO~\cite{shao2024deepseekmath} utilizing the QA-only dataset introduced in Sec.~\ref{sec:data}.

For a given question, the policy $\pi_\theta$ samples a group of $G$ candidate responses $\{y_i\}_{i=1}^{G}$. 
Each response is evaluated using a composite reward function:
\begin{equation}
\label{eq:reward}
R(y_i) = \lambda_{\text{acc}}\,R_{\text{acc}}(y_i) + \lambda_{\text{fmt}}\,R_{\text{fmt}}(y_i),
\end{equation}
where $R_{\text{acc}}, R_{\text{fmt}} \in \{0, 1\}$ reward answer correctness and the ``think with 4D'' format, respectively. 
The group-normalized advantages are then computed as follows:
\begin{equation}
\label{eq:advantage}
\hat{A}_i = \frac{R(y_i) - \mu_G}{\sigma_G}, \quad \mu_G = \tfrac{1}{G}\textstyle\sum_{j} R(y_j),\;\; \sigma_G = \sqrt{\tfrac{1}{G}\textstyle\sum_{j}(R(y_j) - \mu_G)^2}.
\end{equation}

The policy is optimized via a clipped surrogate objective, regularized by the KL divergence against the frozen DIFT reference policy $\pi_{\text{ref}}$.
A key modification over standard GRPO is that we restrict the policy gradient to the index set $\mathcal{T}_{\text{txt}}^{(i)} = \{1, \ldots, |y_i|\} \setminus \mathcal{T}_{\text{lat}}^{(i)}$, which \emph{explicitly excludes all latent token positions}.
This is to \emph{avoid destabilizing gradient noise caused by the mismatch between continuous latent propagation (Eq.~\eqref{eq:autoreg_vis}) and discrete log-probabilities.}
The resulting 4DRL objective is:
\begin{equation}
\label{eq:grpo}
\mathcal{L}_{\text{4DRL}} = -\frac{1}{G}\sum_{i=1}^{G}\frac{1}{|\mathcal{T}_{\text{txt}}^{(i)}|}\sum_{t \,\in\, \mathcal{T}_{\text{txt}}^{(i)}}\bigg[\min\!\Big(\rho_{i,t}\,\hat{A}_i,\;\operatorname{clip}\big(\rho_{i,t},\, 1{-}\epsilon,\, 1{+}\epsilon\big)\hat{A}_i\Big) - \beta\, D_{\text{KL}}^{(t)}\bigg],
\end{equation}
where $\rho_{i,t} = \frac{\pi_\theta(x_t \mid x_{<t})}{\pi_{\text{ref}}(x_t \mid x_{<t})}$ and $D_{\text{KL}}^{(t)}$ are the per-token importance ratio and KL divergence, respectively. 
\definecolor{top1}{HTML}{F4A460}    
\definecolor{top2}{HTML}{FCCC8C}    
\definecolor{top3}{HTML}{FDE8CD}    
\definecolor{rank1}{HTML}{6DBF6D}   
\definecolor{rank2}{HTML}{A5D6A7}   
\definecolor{rank3}{HTML}{D0ECCE}   
\definecolor{gainclr}{RGB}{0,110,55}
\definecolor{secbg}{HTML}{E3F2FD}   

\newcommand{\up}[1]{\textsuperscript{\scriptsize\textcolor{gainclr}{$\uparrow$#1}}}

\begin{table*}[t]
\centering
\small
\setlength{\tabcolsep}{3.6pt}
\renewcommand{\arraystretch}{1.05}
\caption{\textbf{Fine-grained 4D reasoning evaluation on DSR-Bench.}
Top three performers in each column are highlighted from
\colorbox{top1}{Dark} to \colorbox{top3}{Light},
and overall Avg.\ rankings from
\colorbox{rank1}{Dark} to \colorbox{rank3}{Light}.
For 4DThinker, \textit{+DIFT} adds supervised training on top of the base model; \textit{+DIFT+4DRL} further adds RL training. Gains ($\uparrow$) are relative to the base model.}
\vspace{-0.3em}
\label{tab:dsrbench}
\resizebox{\textwidth}{!}{%
\begin{tabular}{l ccccccccccccc c}
\toprule
\textbf{Models}
& \textbf{A.Dis} & \textbf{A.Dir} & \textbf{A.Ori} & \textbf{A.Spd}
& \textbf{A.SpdC} & \textbf{A.DirP}
& \textbf{R.Dis} & \textbf{R.Dir} & \textbf{R.Ori} & \textbf{R.Spd}
& \textbf{R.SpdC} & \textbf{R.DirP}
& \textbf{N-Temp}
& \textbf{Avg.} \\
\midrule
\multicolumn{15}{c}{\textbf{\textit{Proprietary Models}}} \\[1pt]
GPT-5
  & 21.1 & 41.5 & 48.7 & 34.5 & 33.3 & 34.7
  & 17.2 & 44.3 & 41.9 & 21.2 & 25.0 & 30.9
  & 26.7 & \textbf{30.8} \\
Gemini-2.5-Pro
  & 20.0 & 44.6 & 53.6 & 27.3 & 38.7 & 30.5
  & 23.2 & 32.9 & 43.2 & 17.1 & 28.5 & 27.9
  & 34.3 & \textbf{31.7} \\
\midrule
\multicolumn{15}{c}{\textbf{\textit{Spatial Understanding Models}}} \\[1pt]
VLM-3R
  & 28.2 & 27.6 & 31.7 & 42.8 & 38.7 & 33.0
  & 34.4 & 23.8 & 30.8 & 22.2 & 26.4 & 29.1
  & 35.0 & \textbf{31.4} \\
VG-LLM
  & 55.2 & 32.3 & 58.5 & 57.1 & 51.6 & 32.2
  & 56.0 & 36.3 & 32.0 & 30.3 & 32.1 & 29.1
  & 27.9 & \textbf{38.4} \\
DSR Suite-Model
  & \cellcolor{top1}87.0 & \cellcolor{top2}73.8 & \cellcolor{top1}84.1 & \cellcolor{top3}73.8 & \cellcolor{top1}72.0 & 35.5
  & \cellcolor{top1}75.8 & \cellcolor{top2}76.1 & \cellcolor{top2}77.7 & \cellcolor{top3}60.6 & 37.1 & 35.1
  & 46.4 & \cellcolor{rank3}\textbf{58.9} \\
\midrule
\multicolumn{15}{c}{\textbf{\textit{Different Base Models (\method)}}} \\[1pt]

Qwen2.5-VL-3B
  & 20.5 & 20.3 & 20.3 & 37.0 & 17.6 & 22.6
  & 25.5 & 20.7 & 25.6 & 23.7 & 29.0 & 24.2
  & 27.0 & \textbf{24.6} \\
\quad \textit{+DIFT}
  & 48.2 & 28.1 & 25.3 & 46.9 & 28.6 & 28.7
  & 35.8 & 29.9 & 37.2 & 24.7 & 25.3 & 31.1
  & 26.6 & \textbf{31.1}\up{6.5} \\
\quad \textit{+DIFT+4DRL}
  & 50.6 & 34.4 & 29.1 & 56.8 & 34.1 & 34.8
  & 49.1 & 33.3 & 30.8 & 27.8 & 26.8 & 28.6
  & 27.5 & \textbf{34.2}\up{9.6} \\
\addlinespace[4pt]

Qwen2.5-VL-7B
  & 18.8 & 15.3 & 14.6 & 42.8 & 29.0 & 19.4
  & 31.8 & 19.3 & 11.1 & 22.2 & 19.2 & 20.2
  & 30.1 & \textbf{23.5} \\
\quad \textit{+DIFT}
  & 45.8 & 35.9 & 34.2 & 50.6 & 37.4 & 29.6
  & 40.6 & 37.9 & 35.9 & 30.9 & 19.6 & 27.3
  & 38.6 & \textbf{34.8}\up{11.3} \\
\quad \textit{+DIFT+4DRL}
  & 71.1 & 62.5 & 60.8 & 70.4 & 49.5 & 39.1
  & 60.4 & 47.1 & 43.6 & 38.1 & 32.6 & 32.3
  & 42.5 & \textbf{47.1}\up{23.6} \\
\addlinespace[4pt]

Qwen3-VL-8B
  & 23.5 & 24.6 & 42.6 & 29.7 & 27.9 & 33.8
  & 18.1 & 28.4 & 34.5 & 24.2 & 22.1 & 27.9
  & 33.5 & \textbf{28.7} \\
\quad \textit{+DIFT}
  & 51.8 & 45.3 & 55.7 & 44.4 & 37.4 & \cellcolor{top3}42.6
  & 50.0 & 47.1 & 44.9 & 42.3 & 35.5 & 32.9
  & 40.8 & \textbf{42.6}\up{13.9} \\
\quad \textit{+DIFT+4DRL}
  & 78.3 & 65.6 & \cellcolor{top3}82.3 & 59.3 & 58.2 & \cellcolor{top2}46.1
  & 59.4 & 63.2 & 61.5 & 58.8 & 39.1 & \cellcolor{top3}36.6
  & 47.2 & \textbf{54.6}\up{25.9} \\
\addlinespace[4pt]

Qwen3-VL-32B
  & 14.5 & 37.5 & 24.1 & 43.2 & 27.5 & 38.3
  & 13.2 & 32.2 & 38.5 & 17.5 & 20.3 & 28.0
  & 31.8 & \textbf{28.0} \\
\quad \textit{+DIFT}
  & 53.0 & 65.6 & 40.5 & 45.7 & 56.0 & 41.7
  & 50.9 & 48.3 & 46.2 & 43.3 & 37.7 & 34.8
  & 44.2 & \textbf{45.2}\up{17.2} \\
\quad \textit{+DIFT+4DRL}
  & \cellcolor{top3}84.3 & \cellcolor{top1}79.7 & \cellcolor{top3}82.3 & \cellcolor{top1}84.0 & \cellcolor{top2}67.0 & \cellcolor{top1}47.8
  & \cellcolor{top3}67.0 & \cellcolor{top1}82.8 & \cellcolor{top1}79.5 & \cellcolor{top2}62.9 & \cellcolor{top2}44.2 & \cellcolor{top1}41.0
  & \cellcolor{top3}48.5 & \cellcolor{rank1}\textbf{62.0}\up{34.0} \\
\addlinespace[4pt]

InternVL3.5-8B
  & 23.5 & 27.6 & 28.0 & 34.5 & 24.7 & 27.9
  & 22.4 & 17.0 & 19.7 & 28.2 & 30.0 & 14.2
  & 30.1 & \textbf{25.4} \\
\quad \textit{+DIFT}
  & 43.4 & 42.2 & 36.7 & 37.0 & 33.0 & 30.4
  & 41.5 & 39.1 & 37.2 & 35.1 & 30.4 & 28.0
  & 41.2 & \textbf{36.2}\up{10.8} \\
\quad \textit{+DIFT+4DRL}
  & 69.9 & 64.1 & 65.8 & 53.1 & 51.6 & 39.1
  & 52.8 & 55.2 & 52.6 & 47.4 & 39.1 & \cellcolor{top3}36.6
  & \cellcolor{top1}49.8 & \textbf{50.0}\up{24.6} \\
\addlinespace[4pt]

InternVL3.5-38B
  & 25.8 & 27.8 & 29.2 & 34.2 & 24.7 & 28.5
  & 26.7 & 16.3 & 23.4 & 29.2 & 32.1 & 15.4
  & 31.3 & \textbf{26.7} \\
\quad \textit{+DIFT}
  & 59.0 & 53.1 & 49.4 & 40.7 & 41.8 & 32.2
  & 51.9 & 41.4 & 38.5 & 44.3 & \cellcolor{top3}39.9 & 29.8
  & 44.2 & \textbf{42.5}\up{15.8} \\
\quad \textit{+DIFT+4DRL}
  & \cellcolor{top2}85.5 & \cellcolor{top3}73.4 & \cellcolor{top2}83.5 & \cellcolor{top2}80.2 & \cellcolor{top3}60.4 & 37.4
  & \cellcolor{top2}67.9 & \cellcolor{top3}74.7 & \cellcolor{top3}70.5 & \cellcolor{top1}64.9 & \cellcolor{top1}44.9 & \cellcolor{top2}37.3
  & \cellcolor{top1}49.8 & \cellcolor{rank2}\textbf{59.4}\up{32.7} \\
\bottomrule
\end{tabular}
}%
\vspace{-1.2em}
\end{table*}

\section{Experiments}
\label{sec:exp}
\paragraph{Experimental setup.}
We follow the two-stage training pipeline described in Sec.~\ref{sec:training}. Implementation details are provided in Appendix~\ref{app:implementation}, training data composition and evaluation benchmarks in Appendix~\ref{app:datasets}, and subtask definitions in Appendix~\ref{app:subtasks}. All benchmarks are formatted as multiple-choice questions; we report accuracy via exact match, with ``Avg.’’ denoting the mean across all subtasks.

\paragraph{Baselines.}
We compare with three groups of models: (1) \emph{proprietary VLMs}: GPT-5~\cite{gpt5} and Gemini-2.5-Pro~\cite{gemini2.5}; (2) \emph{spatial understanding models}: VLM-3R~\cite{vlm3r}, VG-LLM~\cite{vg-llm}, DSR Suite-Model~\cite{dsrsuite}, SpaceR-7B~\cite{ouyang2025spacer}, VST-7B-RL~\cite{vst}, Spatial-SSRL-7B~\cite{liu2025spatialssrl}, and SpatialLadder-3B~\cite{li2025spatialladder}; and (3) \emph{base VLMs} on which \method is applied: Qwen2.5-VL-3B/7B~\cite{bai2025qwen2.5}, Qwen3-VL-8B/32B~\cite{qwen3technicalreport}, and InternVL3.5-8B/38B~\cite{wang2025internvl3_5}.

\subsection{Benchmarking Fine-Grained 4D Reasoning}
\label{sec:dsrbench}
DSR-Bench~\cite{dsrsuite} targets \emph{quantitative geometric measurement} in dynamic scenes, requiring models to produce procedural answers that precisely characterize how spatial attributes (e.g., distance, orientation, speed) evolve over time.


As shown in Tab.~\ref{tab:dsrbench}, \method delivers consistent improvements across all base VLMs. For Qwen2.5-VL-3B, DIFT yields \textbf{+6.5 pp} (31.1 vs.\ 24.6), and the full \method pipeline (DIFT+4DRL) achieves \textbf{+9.6 pp} (34.2 vs.\ 24.6). The gains become more pronounced at larger scales, i.e., Qwen3-VL-32B improves by \textbf{+34.0 pp} (\textbf{62.0} vs.\ 28.0), \emph{surpassing both the proprietary Gemini-2.5-Pro (31.7) and the best task-specific DSR Suite-Model (58.9)}. Cross-architecture generalization is also evident, with InternVL3.5-38B gaining \textbf{+32.7 pp} (59.4 vs.\ 26.7).

Notably, the gains are most pronounced on absolute subtasks (e.g., A.Dis, A.Ori) where base models near chance (${\sim}$20\%), indicating that \emph{4D latents supply the geometric grounding that pure language reasoning lacks.} Moreover, our best model surpasses previous state-of-the-art (SOTA) DSR Suite-Model (\textbf{62.0} vs.\ 58.9) without external 3D priors (e.g., Geometry Selection Module (GSM)), showing that \emph{internalized 4D imagery can be more effective than modular geometric injection.}

\subsection{Comparison on Holistic Dynamic Understanding}
\label{sec:dynbench}


\definecolor{top1}{HTML}{F4A460}    
\definecolor{top2}{HTML}{FCCC8C}    
\definecolor{top3}{HTML}{FDE8CD}    
\definecolor{rank1}{HTML}{6DBF6D}   
\definecolor{rank2}{HTML}{A5D6A7}   
\definecolor{rank3}{HTML}{D0ECCE}   
\definecolor{gainclr}{RGB}{0,110,55}
\definecolor{secbg}{HTML}{E3F2FD}   

\begin{table*}[t]
\centering
\noindent
\begin{minipage}[t]{0.63\textwidth}
\centering
\scriptsize
\setlength{\tabcolsep}{3pt}
\renewcommand{\arraystretch}{1.1}
\captionof{table}{\textbf{Holistic dynamic understanding evaluation on Dyn-Bench}. Top three performers in each column are highlighted.}
\vspace{-0.3em}
\label{tab:dynbench}
\begin{tabular*}{\linewidth}{@{\extracolsep{\fill}}l cccc}
\toprule
\textbf{Methods} & \textbf{Inter-Obj} & \textbf{Obj-Scene} & \textbf{Cam-Obj} & \textbf{Avg.} \\
\midrule
\multicolumn{5}{c}{\textbf{\textit{Proprietary Models}}} \\[1pt]
GPT-5~\cite{gpt5}            & 54.7 & 70.2 & 59.2 & \textbf{61.4} \\
Gemini-2.5-Pro~\cite{gemini2.5}   & 56.1 & 64.4 & 55.8 & \textbf{58.8} \\
\midrule
\multicolumn{5}{c}{\textbf{\textit{Spatial Understanding Models}}} \\[1pt]
SpaceR-7B~\cite{ouyang2025spacer}        & 56.2 & 72.7 & 48.6 & \textbf{59.2} \\
VST-7B-RL~\cite{vst}        & 56.3 & 74.4 & 45.7 & \textbf{58.8} \\
Spatial-SSRL-7B~\cite{liu2025spatialssrl}  & 47.5 & 69.4 & 36.7 & \textbf{51.2} \\
SpatialLadder-3B~\cite{li2025spatialladder} & 52.0 & 72.7 & 44.0 & \textbf{56.2} \\
\midrule
\multicolumn{5}{c}{\textbf{\textit{Different Base Models (\method)}}} \\[1pt]
Qwen2.5-VL-7B~\cite{bai2025qwen2.5}        & 50.8 & 69.9 & 42.1 & \textbf{54.3} \\
\quad \textit{+DIFT}          & 56.3 & 74.4 & 47.8 & \textbf{59.5}\up{5.2} \\
\quad \textit{+DIFT+4DRL}     & 63.5 & 79.2 & 54.9 & \textbf{65.9}\up{11.6} \\
\addlinespace[3pt]
Qwen3-VL-8B~\cite{qwen3technicalreport}          & 59.0 & 76.2 & 56.0 & \textbf{63.7} \\
\quad \textit{+DIFT}          & 64.1 & 79.0 & 60.8 & \textbf{68.0}\up{4.3} \\
\quad \textit{+DIFT+4DRL}     & \cellcolor{top2}70.9 & \cellcolor{top1}84.5 & \cellcolor{top2}68.0 & \cellcolor{rank2}\textbf{74.5}\up{10.8} \\
\addlinespace[3pt]
Qwen3-VL-32B~\cite{qwen3technicalreport}         & 61.1 & 76.0 & 56.5 & \textbf{64.5} \\
\quad \textit{+DIFT}          & \cellcolor{top3}66.3 & \cellcolor{top3}79.8 & \cellcolor{top3}61.5 & \cellcolor{rank3}\textbf{69.2}\up{4.7} \\
\quad \textit{+DIFT+4DRL}     & \cellcolor{top1}73.6 & \cellcolor{top2}84.0 & \cellcolor{top1}68.7 & \cellcolor{rank1}\textbf{75.4}\up{10.9} \\
\addlinespace[3pt]
InternVL3.5-8B~\cite{wang2025internvl3_5}        & 50.8 & 65.3 & 42.7 & \textbf{52.9} \\
\quad \textit{+DIFT}          & 56.3 & 69.8 & 48.2 & \textbf{58.1}\up{5.2} \\
\quad \textit{+DIFT+4DRL}     & 63.1 & 75.5 & 54.9 & \textbf{64.5}\up{11.6} \\
\addlinespace[3pt]
InternVL3.5-38B~\cite{wang2025internvl3_5}       & 49.4 & 64.9 & 44.3 & \textbf{52.9} \\
\quad \textit{+DIFT}          & 55.4 & 70.1 & 50.3 & \textbf{58.6}\up{5.7} \\
\quad \textit{+DIFT+4DRL}     & 62.9 & 76.6 & 57.9 & \textbf{65.8}\up{12.9} \\
\bottomrule
\end{tabular*}
\end{minipage}%
\hfill%
\begin{minipage}[t]{0.33\textwidth}
\raggedright
\scriptsize
\setlength{\tabcolsep}{4pt}
\renewcommand{\arraystretch}{1.2}
\captionsetup{justification=raggedright,singlelinecheck=false}
\captionof{table}{Comparison of different VLM training strategies.}
\label{tab:train_strategy}
\vspace{-0.3em}
\begin{tabular*}{\linewidth}{@{\extracolsep{\fill}}lc}
\toprule
\textbf{Training strategies} & \textbf{Accuracy} \\
\midrule
Raw QA SFT       & 25.1 \\
CoT SFT          & 26.8 \\
DIFT             & \textbf{31.1} \\
\midrule
GRPO             & 27.1 \\
CoT SFT + GRPO   & 29.7 \\
DIFT + 4DRL      & \textbf{34.2} \\
\bottomrule
\end{tabular*}

\vspace{10.7pt}
\captionof{table}{Ablation on different loss and reward components.}
\label{tab:loss_reward}
\vspace{1.9pt}
\begin{tabular*}{\linewidth}{@{\extracolsep{\fill}}lc}
\toprule
\textbf{Loss \& reward components} & \textbf{Accuracy} \\
\midrule
Full  & \textbf{34.2} \\
\midrule
w/o $\mathcal{L}_{\text{sim}}$   & 28.5 \\
w/o $\mathcal{L}_{\text{ce}}$    & 19.3 \\
w/o $R_{\text{fmt}}$             & 33.4 \\
w/o $R_{\text{acc}}$             & 32.0 \\
\bottomrule
\end{tabular*}

\vspace{10.7pt}
\captionof{table}{Effect of different latent token sizes for DIFT training.}
\label{tab:latent_size}
\vspace{1.9pt}
\begin{tabular*}{\linewidth}{@{\extracolsep{\fill}}lccccc}
\toprule
\textbf{Size} & 1 & 2 & 4 & 8 & 16 \\
\midrule
\textbf{Accuracy} & 29.6 & 30.4 & \textbf{31.1} & 30.8 & 29.3 \\
\bottomrule
\end{tabular*}
\end{minipage}
\vspace{-1.4em}
\end{table*}

Beyond geometric precision, Dyn-Bench~\cite{thinkdynamic} evaluates \emph{semantic-level} dynamic understanding along three axes: \emph{inter-object}, \emph{object-scene}, and \emph{camera-object}, testing whether models can reason about interactions, trajectories, and causal relationships in 4D environments.

Tab.~\ref{tab:dynbench} shows that \method yields strong and consistent gains, \emph{with the top three performers on the leaderboard all produced by our method.}
In terms of Qwen2.5-VL-7B, DIFT+4DRL achieves \textbf{+11.6 pp} (65.9 vs.\ 54.3), already outperforming GPT-5 (61.4) and Gemini-2.5-Pro (58.8), while Qwen3-VL-32B reaches \textbf{75.4} (\textbf{+10.9 pp}), establishing a new SOTA.
Even with the smaller Qwen2.5-VL-7B backbone, our model surpasses all dedicated spatial understanding baselines, including SpaceR-7B (59.2), VST-7B-RL (58.8), and SpatialLadder-3B (56.2), by a substantial margin.

The strong performance on Dyn-Bench reveals that the 4D latents learned by \method encode not only low-level geometric trajectories (as validated of DSR-Bench on Tab.~\ref{tab:dsrbench}), but also \emph{higher-level motion semantics}. Notably, improvements on the camera-object axis, \emph{which demands disentangling ego-motion from object dynamics, confirm that \method acquires a genuine internal 4D representation rather than relying on single-viewpoint heuristics.}

\subsection{Ablation Studies}
\label{sec:ablation}

We further ablate key design choices of \method using Qwen2.5-VL-3B on DSR-Bench.

\paragraph{Training strategy.}
As shown in Tab.~\ref{tab:train_strategy}, under supervised training, Raw QA SFT (25.1) and CoT SFT (26.8) barely exceed the base model, while DIFT reaches \textbf{31.1} (\textbf{+5.2 pp}) by jointly supervising latent visual tokens and textual reasoning \emph{in a unified ``think with 4D'' paradigm}. Under reinforced training, vanilla GRPO (27.1) and CoT SFT + GRPO (29.7) \emph{remain limited by the expressiveness bottleneck of discrete text}, whereas DIFT + 4DRL achieves \textbf{34.2}, confirming that 4D latent representations offer a richer optimization landscape than discrete textual CoT.

\paragraph{Loss and reward components.}
Tab.~\ref{tab:loss_reward} reveals the relative importance of each component. Removing $\mathcal{L}_{\text{ce}}$ causes catastrophic degradation (34.2 $\to$ 19.3), as the model \emph{loses the ability to generate coherent text}. Removing $\mathcal{L}_{\text{sim}}$ yields the next largest drop (34.2 $\to$ 28.5), confirming that visual alignment supervision is essential, i.e., \emph{4D latents degenerate into ungrounded representations without it.} For RL rewards, $R_{\text{acc}}$ (34.2 $\to$ 32.0) contributes more than $R_{\text{fmt}}$ (34.2 $\to$ 33.4), \emph{as accuracy reward directly shapes reasoning quality} while format reward mainly ensures structural compliance.

\paragraph{Latent token size.}
Tab.~\ref{tab:latent_size} studies the capacity-performance trade-off. Accuracy increases from $K{=}1$ (29.6) to $K{=}4$ (31.1), as \emph{additional tokens encode richer spatial-temporal information for imagery.} Beyond $K{=}4$, performance slightly declines (30.8 at $K{=}8$, 29.3 at $K{=}16$). We attribute this to \emph{excessive latent tokens diluting the textual context}, disrupting the model's language coherence.

\section{Conclusion and Limitation}
\label{sec:conclusion}

In this paper, we propose the ``think with 4D'' framework \method, 
which enables VLMs to reason about dynamic scenes through latent visual imagery.
To this end, we integrate a scalable and annotation-free data generation pipeline, joint text-imagery supervision fine-tuning(DIFT), and outcome-based 4D reinforcement learning (4DRL) into a unified training recipe.
Consistent improvements across multiple benchmarks confirm that 
grounding chain-of-thought in continuous 4D imagery is more effective than purely textual or module-augmented approaches.

\paragraph{Limitation \& Future Work.}
While \method demonstrates consistent improvements, we recognize the following limitations:
(1) The current data pipeline relies on off-the-shelf geometric estimators (e.g., MegaSaM), whose errors may propagate into the training data. Although our 4DRL partially mitigates such noise, incorporating more robust geometric priors could further improve data quality.
(2) Our evaluation focuses on multiple-choice benchmarks for dynamic reasoning; extending the framework to open-ended generation tasks (e.g., embodied planning) remains a challenge.


{
    \small
    \bibliographystyle{ieeenat_fullname}
    \bibliography{main}
}

\newpage

\appendix

\section{Object Selection Rules}
\label{app:rules}

As described in Sec.~\ref{sec:data}, we define a set of predefined rules $\mathcal{R}$ to guide $M_{\text{high}}$ in selecting a representative static object $o^s$ and a dynamic object $o^d$ from each video. Specifically, we instruct $M_{\text{high}}$ with the following criteria:

\paragraph{Static object selection.}
\begin{itemize}[nosep,leftmargin=1.5em]
    \item The object must be \emph{stationary} throughout the entire video (e.g., the traffic sign).
    \item It should be \emph{visually salient} and occupy a reasonable area, avoiding small/occluded objects.
    \item The object should \emph{persist} in most frames without prolonged absence.
    \item Select \emph{central} objects to maximize camera-induced apparent displacement.
\end{itemize}

\paragraph{Dynamic object selection.}
\begin{itemize}[nosep,leftmargin=1.5em]
    \item The object must exhibit \emph{clear, non-trivial} motion (e.g., the walking person wearing a red shirt).
    \item It should be visually \emph{distinguishable} from the background and other moving entities.
    \item The object must be \emph{trackable} across a sufficient number of frames ($\geq$ 50\% of the video duration).
    \item Prefer the most \emph{prominent} moving object if multiple candidates exist.
\end{itemize}

Both selections are formatted as structured outputs containing the object name and a brief visual description (e.g., ``the red car on the left lane''), which are subsequently used as text prompts for SAM3~\cite{sam3} to generate mask sequences.

\section{Prompt Details}
\label{app:prompts}

Our data generation pipeline employs a series of carefully designed prompts at each stage. We present them below, organized by pipeline stage described in Sec.~\ref{sec:data}.

\begin{figure*}[htbp]
    \centering
    \includegraphics[width=\textwidth]{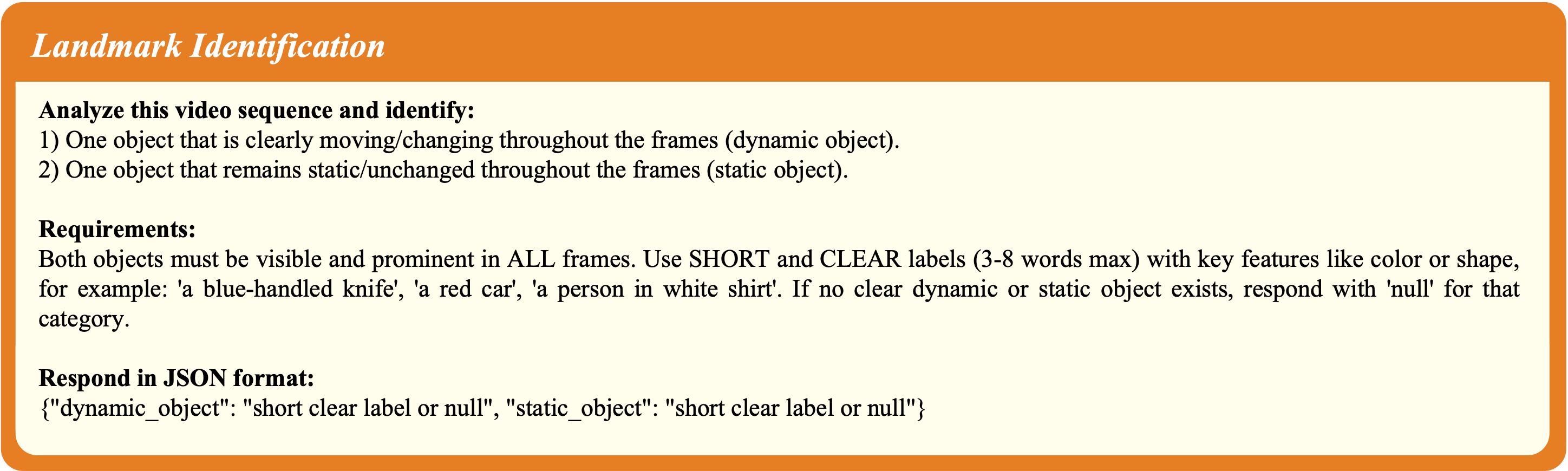}
    \vspace{-0.8em}
    \caption{Prompt for \textbf{landmark identification}. $M_{\text{high}}$ identifies one static and one dynamic object with short visual descriptions, which are subsequently used as text prompts for SAM3 mask extraction.}
    \label{fig:prompt_landmark}
\end{figure*}

\begin{figure*}[htbp]
    \centering
    \includegraphics[width=\textwidth]{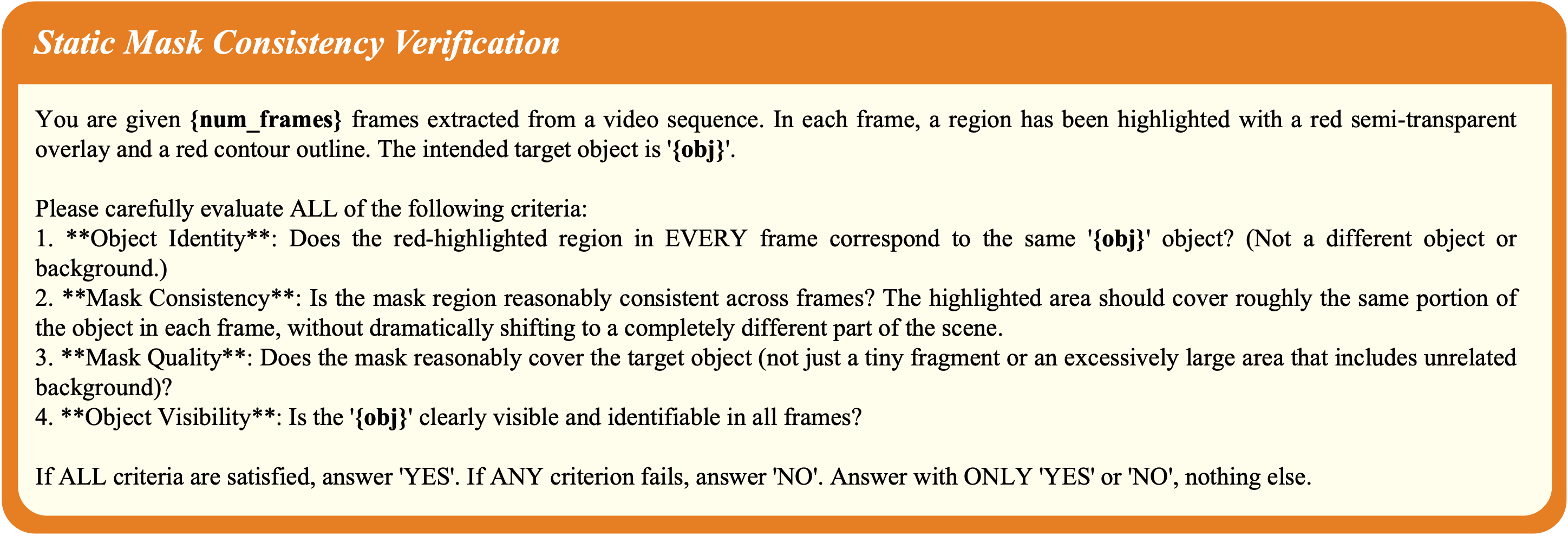}
    \vspace{-0.8em}
    \caption{Prompt for \textbf{static mask consistency verification}. $M_{\text{high}}$ evaluates four criteria across all overlay frames to implement the consistency filter (Eq.~\eqref{eq:filter}).}
    \label{fig:prompt_static_verify}
\end{figure*}

\begin{figure*}[htbp]
    \centering
    \includegraphics[width=\textwidth]{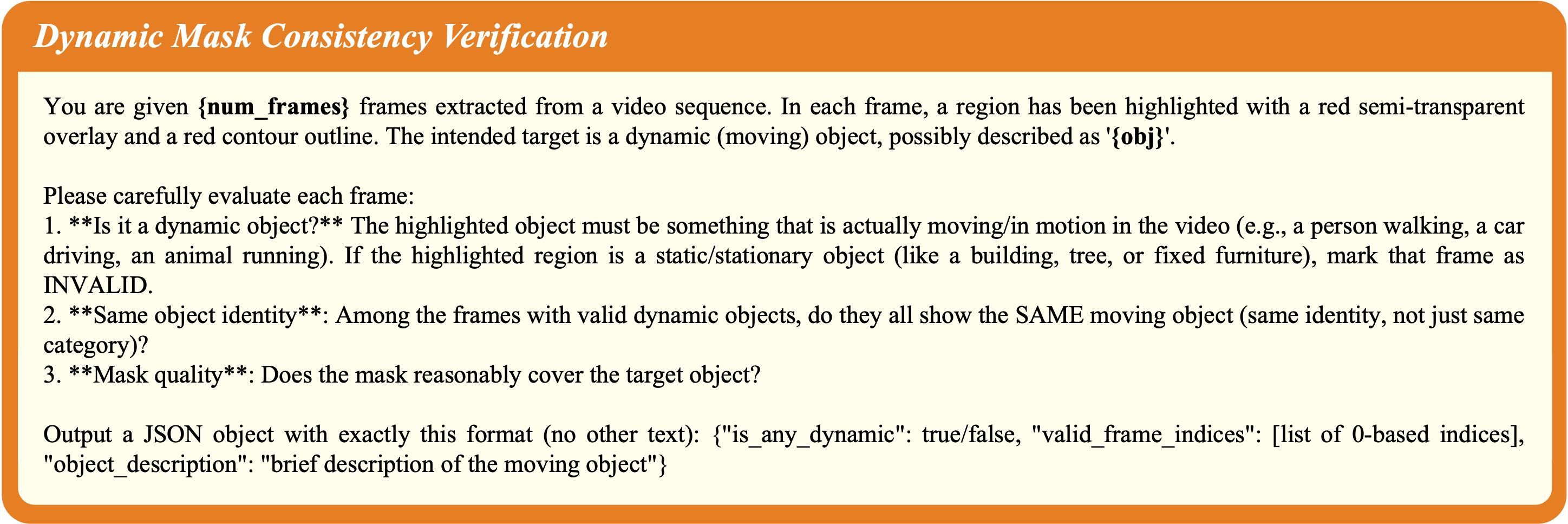}
    \vspace{-0.8em}
    \caption{Prompt for \textbf{dynamic object mask verification}. Unlike the binary static check (Fig.~\ref{fig:prompt_static_verify}), this prompt returns per-frame validity indices, allowing partial acceptance of frames.}
    \label{fig:prompt_dynamic_verify}
\end{figure*}

\paragraph{Video preprocessing prompts.}
The preprocessing stage uses three prompts. The \emph{landmark identification} prompt (Fig.~\ref{fig:prompt_landmark}) with rules in Sec.~\ref{app:rules} instructs $M_{\text{high}}$ to select one static and one dynamic object from uniformly sampled frames, producing structured JSON labels for SAM3 mask extraction. The \emph{static mask consistency verification} prompt (Fig.~\ref{fig:prompt_static_verify}) implements the consistency filter $\Phi_{M_{\text{high}}}$ in Eq.~\eqref{eq:filter} by evaluating object identity, mask consistency, mask quality, and visibility across all overlay frames. For dynamic objects, the \emph{dynamic mask verification} prompt (Fig.~\ref{fig:prompt_dynamic_verify}) performs a more nuanced per-frame evaluation, returning \emph{validity indices to allow partial frame acceptance.}

\begin{figure*}[t]
    \centering
    \includegraphics[width=\textwidth]{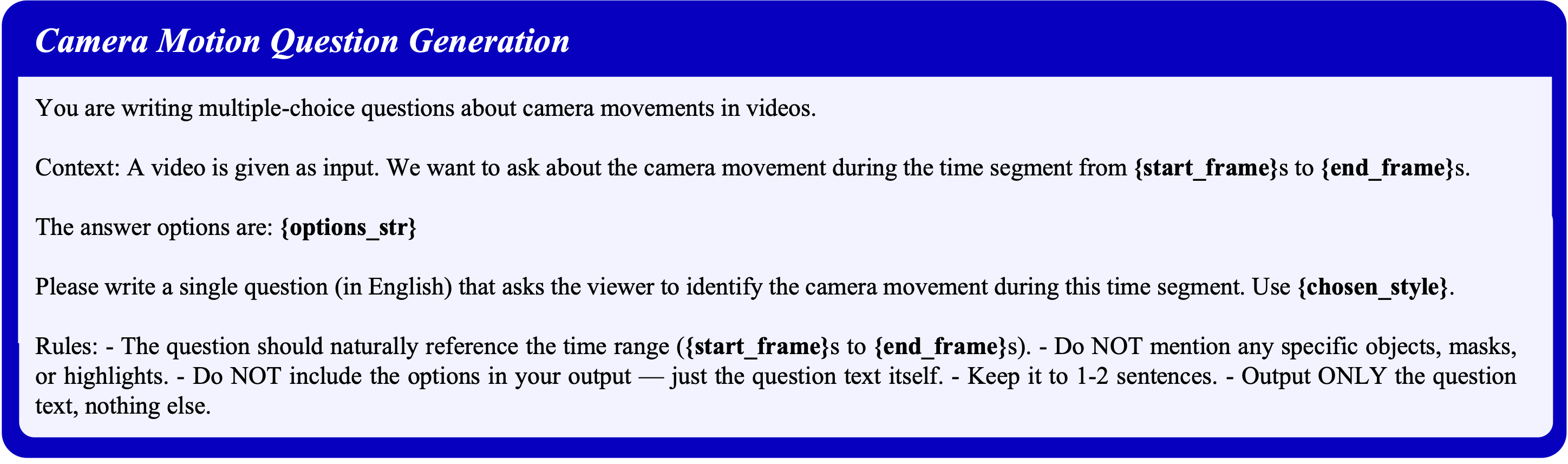}
    \vspace{-0.8em}
    \caption{Prompt for \textbf{camera motion question generation}. Given a time segment and answer options, $M_{\text{high}}$ produces a natural-language MCQ.}
    \label{fig:prompt_camera_q}
\end{figure*}

\paragraph{Camera motion QA prompts.}
For camera motion data, the \emph{question generation} prompt (Fig.~\ref{fig:prompt_camera_q}) produces natural-language MCQs subject to predefined constraints.

\begin{figure*}[t]
    \centering
    \includegraphics[width=\textwidth]{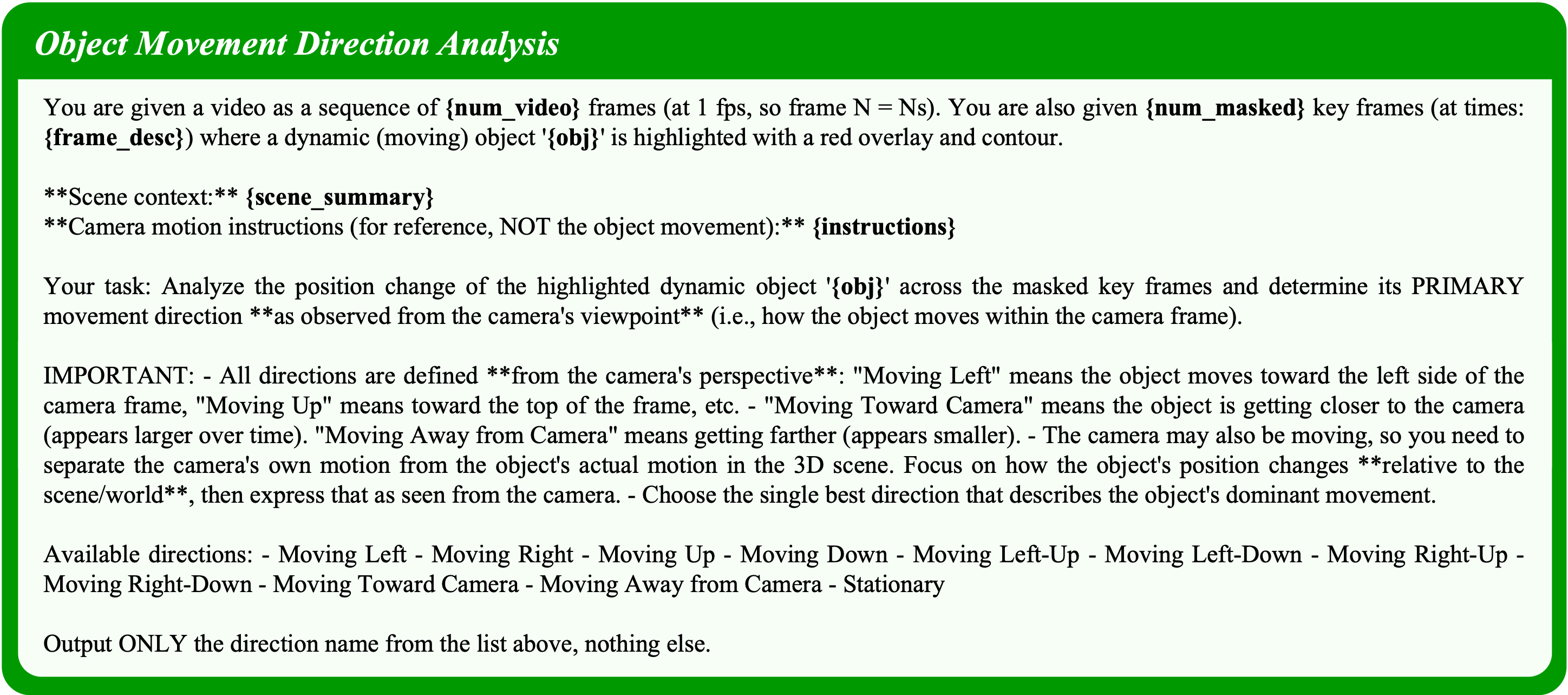}
    \vspace{-0.8em}
    \caption{Prompt for \textbf{object movement direction analysis}. $M_{\text{high}}$ analyzes centroid displacement and apparent scale variation across masked key frames to determine the primary movement direction, while explicitly separating camera ego-motion from the object's own motion.}
    \label{fig:prompt_direction}
\end{figure*}

\begin{figure*}[t]
    \centering
    \includegraphics[width=\textwidth]{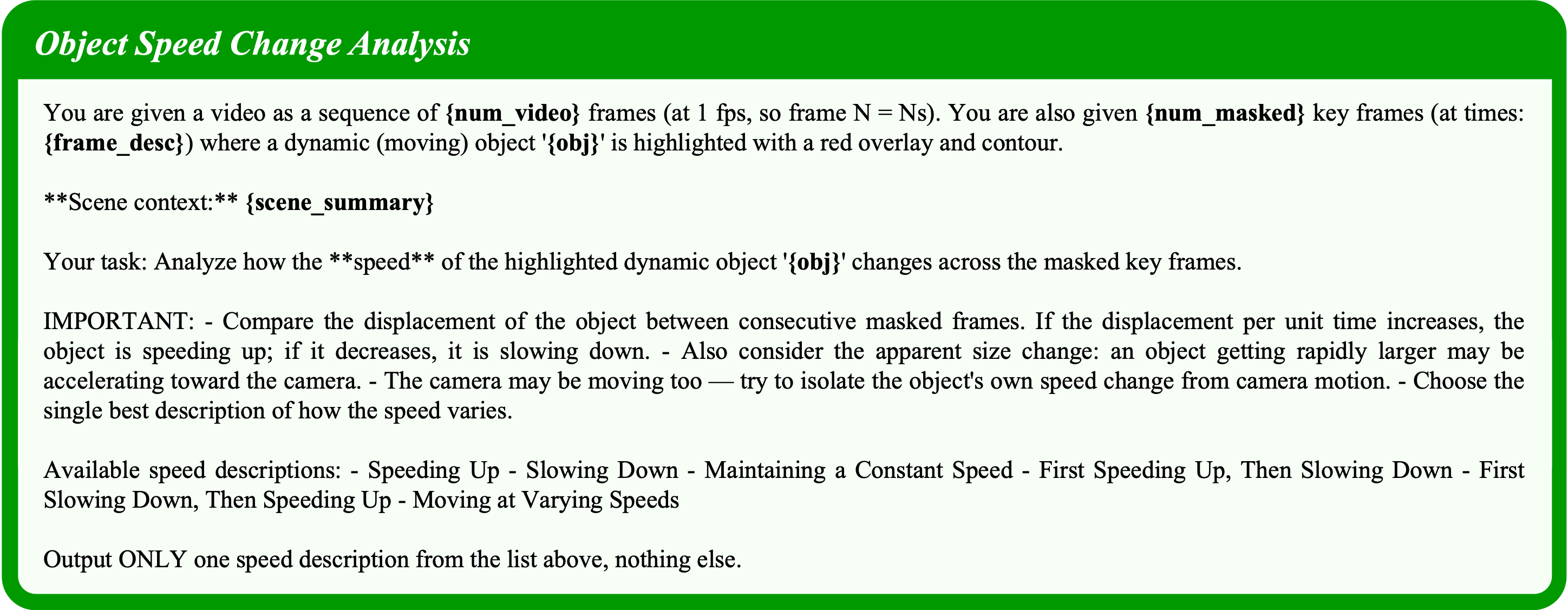}
    \vspace{-0.8em}
    \caption{Prompt for \textbf{object speed change analysis}. Complementary to the Fig.~\ref{fig:prompt_direction}, $M_{\text{high}}$ classifies the speed pattern by comparing per-frame centroid displacements and apparent size changes.}
    \label{fig:prompt_speed}
\end{figure*}

\begin{figure*}[t]
    \centering
    \includegraphics[width=\textwidth]{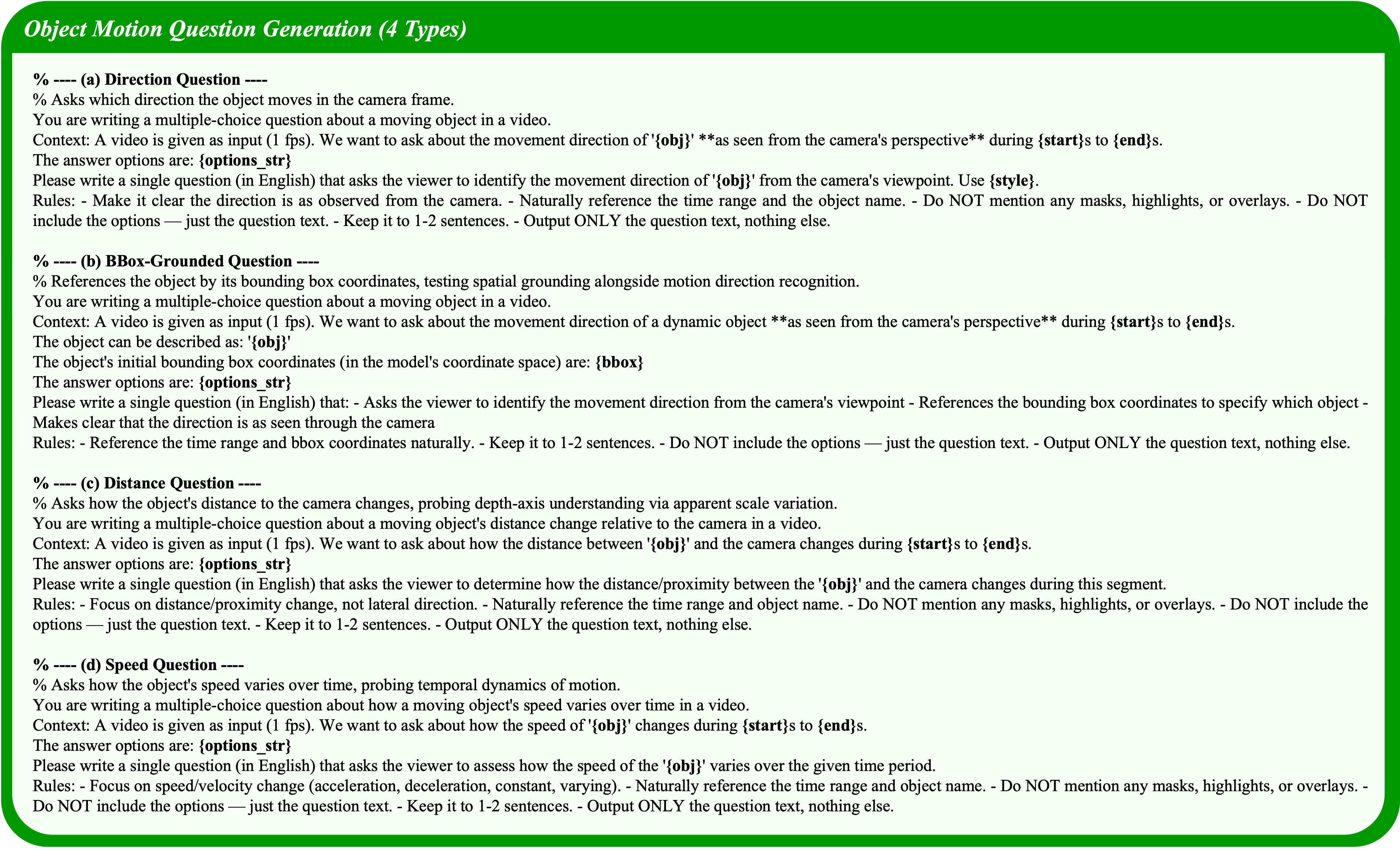}
    \vspace{-0.8em}
    \caption{Prompts for \textbf{object motion question generation} (four types). Each variant probes a different aspect of dynamic understanding: (a) movement direction, (b) 4D question with bounding-box grounding, (c) distance change relative to the camera, and (d) speed variation over time.}
    \label{fig:prompt_object_q}
\end{figure*}

\paragraph{Object motion QA prompts.}
Object motion data construction involves three prompt types. First, the trajectory analysis determines ground-truth motion attributes: the \emph{direction analysis} prompt (Fig.~\ref{fig:prompt_direction}) identifies the primary movement direction from 11 candidates (Tab.~\ref{tab:answer_choices}) while separating camera ego-motion, and the \emph{speed change analysis} prompt (Fig.~\ref{fig:prompt_speed}) classifies the speed pattern. Second, the \emph{question generation} prompt (Fig.~\ref{fig:prompt_object_q}) covers four complementary question types: direction, bounding-box grounded description, distance change, and speed variation detailed in Tab.~\ref{tab:question_types}.

\begin{figure*}[t]
    \centering
    \includegraphics[width=\textwidth]{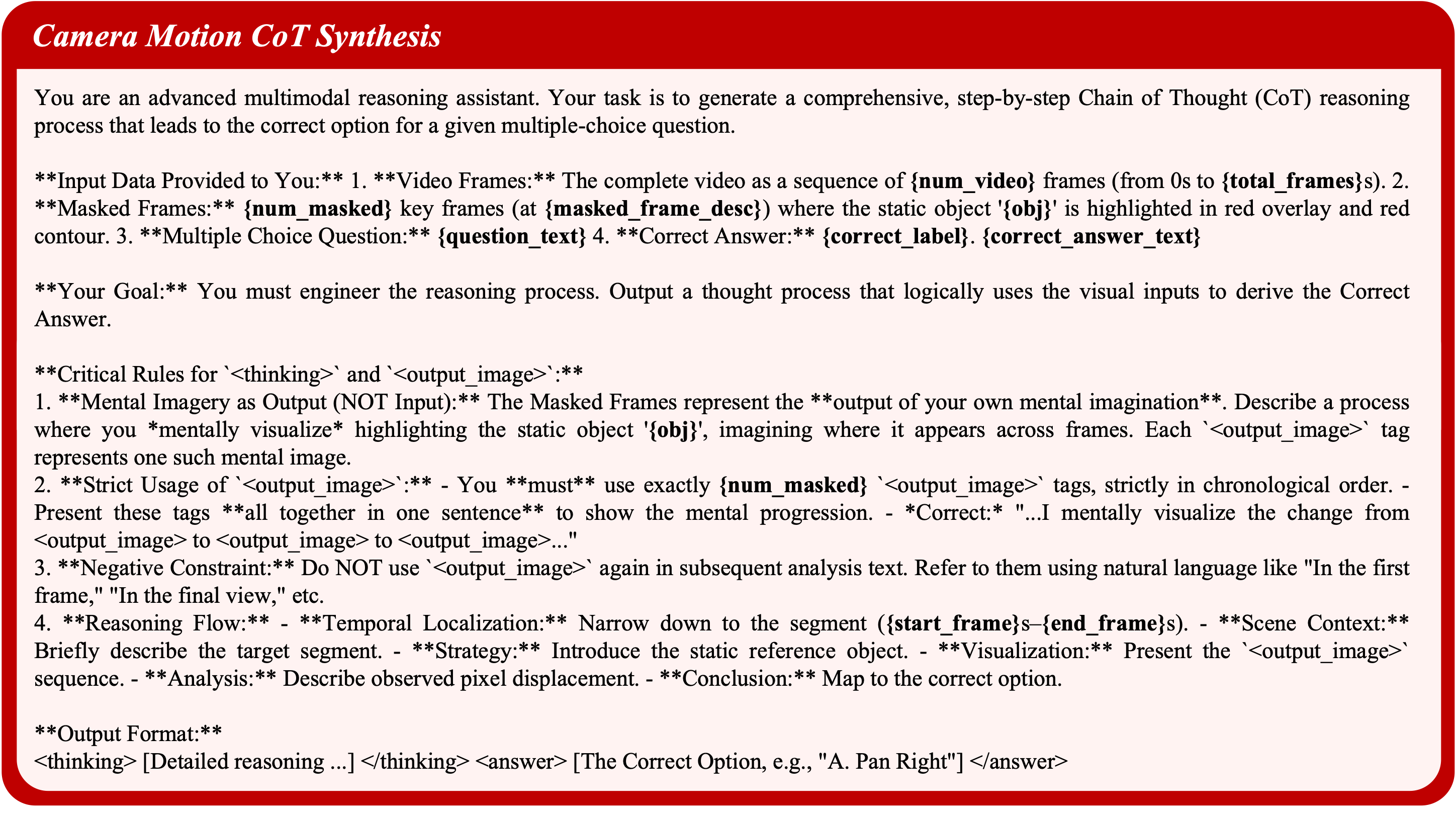}
    \vspace{-0.8em}
    \caption{Prompt for \textbf{camera motion CoT synthesis}. Given the video, static mask overlays, and the correct answer, $M_{\text{high}}$ produces reasoning trace where \texttt{<output\_image>} placeholders represent the model's own ``mental imagery,'' which are later replaced by latent visual tokens during DIFT training.}
    \label{fig:prompt_camera_cot}
\end{figure*}

\begin{figure*}[htbp]
    \centering
    \includegraphics[width=\textwidth]{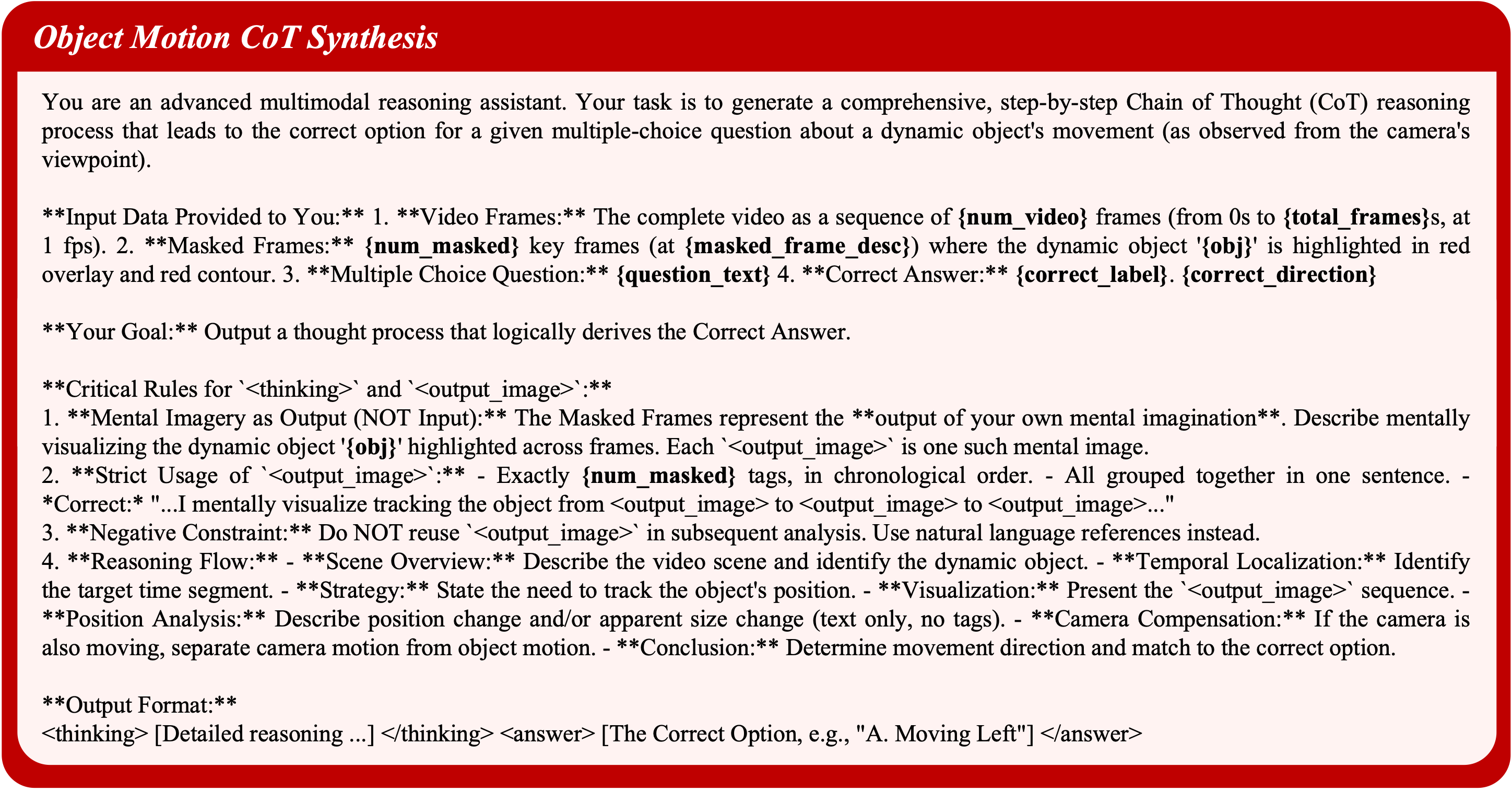}
    \vspace{-0.8em}
    \caption{Prompt for \textbf{object motion CoT synthesis}. Analogous to the camera motion variant (Fig.~\ref{fig:prompt_camera_cot}), the model tracks the dynamic object's position across frames.}
    \label{fig:prompt_object_cot}
\end{figure*}

\paragraph{CoT synthesis prompts.}
The core ``think with 4D'' reasoning traces are produced by the \emph{camera motion CoT synthesis} prompt (Fig.~\ref{fig:prompt_camera_cot}) and the \emph{object motion CoT synthesis} prompt (Fig.~\ref{fig:prompt_object_cot}). Both enforce the \method reasoning flow and frame the overlay images as the model's own mental imagination via placeholders. The object motion variant additionally includes a \emph{camera compensation} step that disentangles camera-induced apparent motion from the object's own displacement.

\begin{figure}[htbp]
    \centering
    \includegraphics[width=\columnwidth]{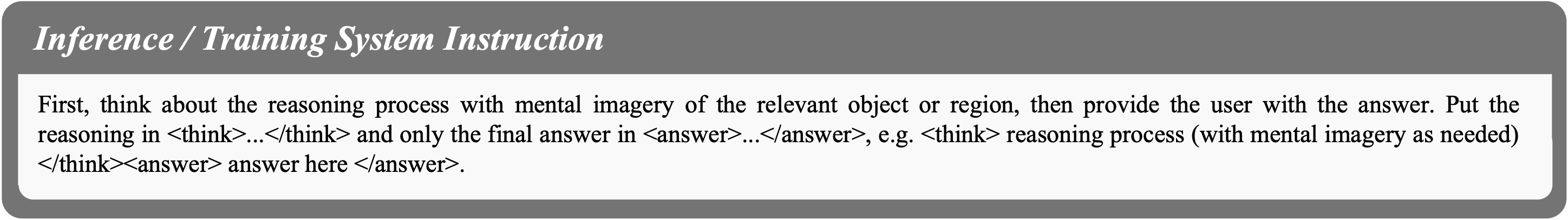}
    \vspace{-0.8em}
    \caption{The \textbf{system instruction} appended before every question during DIFT training, 4DRL training, and inference. It specifies the output format that the model must follow.}
    \label{fig:prompt_system}
\end{figure}

\paragraph{Training and inference instruction.}
Finally, the \emph{system prompt} (Fig.~\ref{fig:prompt_system}) is appended before every question during DIFT training, 4DRL, and inference, specifying the output format that the model must follow. During 4DRL, the format reward $R_{\text{fmt}}$ checks adherence to this think-answer structure.

\begin{table*}[htbp]
\centering
\caption{Candidate question types, target objects, and descriptions in our data generation pipeline.}
\label{tab:question_types}
\vspace{0.3em}
\small
\setlength{\tabcolsep}{5pt}
\begin{tabular}{p{2.0cm}p{2.2cm}p{2cm}p{6.2cm}}
\toprule
\textbf{Category} & \textbf{Type} & \textbf{Target Object} & \textbf{Description} \\
\midrule
Camera Motion & Direction & Static object & Identifying how the camera moves \\
\midrule
\multirow{7}{*}{Object Motion}
  & Direction & Dynamic object & Detecting motion direction from camera view \\
  & BBox-Grounded & Dynamic object & Identifying direction with bbox-based grounding \\
  & Distance & Dynamic object & Determining the object's distance to the camera \\
  & Speed & Dynamic object & Assessing how the object's speed varies over time \\
\bottomrule
\end{tabular}
\end{table*}

\begin{table*}[htbp]
\centering
\caption{Answer choices for each question type.}
\label{tab:answer_choices}
\vspace{0.3em}
\small
\setlength{\tabcolsep}{5pt}
\begin{tabular}{p{2.2cm}p{11cm}}
\toprule
\textbf{Type} & \textbf{Answer Choices} \\
\midrule
Camera Direction & Move Forward, Move Backward, Move Left, Move Right, Move Up, Move Down, Pan Left, Pan Right, Tilt Up, Tilt Down, Roll Clockwise, Roll Counter-clockwise \\
\midrule
Object Direction & Moving Left, Right, Up, Down, Left-Up, Left-Down, Right-Up, Right-Down, Toward Camera, Away from Camera, Stationary \\
\midrule
Object Distance & Moving Toward Camera, Moving Away from Camera, Maintaining Roughly the Same Distance \\
\midrule
Object Speed & Speeding Up, Slowing Down, Maintaining a Constant Speed, First Speeding Up Then Slowing Down, First Slowing Down Then Speeding Up, Moving at Varying Speeds \\
\bottomrule
\end{tabular}
\end{table*}

\section{Candidate Question Types and Answer Choices}
\label{app:question_types}

Tab.~\ref{tab:question_types} summarizes the five \emph{candidate question types} in our data generation pipeline, and Tab.~\ref{tab:answer_choices} lists the \emph{corresponding answer choices}. 
To ensure rigorous evaluation, the camera motion MCQs are structured with four options, comprising the ground-truth label and three plausible distractors. These distractors are preferentially drawn from semantically related motion types to prevent trivial guessing. Following a parallel design, the object motion MCQs maintain the same four-option format, with distractors systematically sampled from the valid candidate set of the respective question category.

\section{Training and Evaluation Datasets}
\label{app:datasets}


\paragraph{Training data.}
The DIFT stage uses $\sim$38K samples synthesized by our pipeline from SpatialVID~\cite{spatialvid}, a large-scale video collection of 2.7M clips (7,089 hours) with MegaSaM-estimated camera poses. We use only its raw videos and geometric annotations, no human labels are involved. The 4DRL stage uses DSR-Train~\cite{dsrsuite}, which provides $\sim$37K QA pairs covering compound camera-object motions from in-the-wild videos, without explicit reasoning traces.

\paragraph{Benchmarks.}
We evaluate on benchmarks spanning dynamic spatial reasoning. \textbf{DSR-Bench}~\cite{dsrsuite} covers 13 fine-grained subtasks including absolute/relative direction, distance, speed, and orientation. \textbf{Dyn-Bench}~\cite{thinkdynamic} evaluates perception, tracking, and reasoning of dynamic content in 4D scenes with 1K videos and 7K VQA pairs. We do not use its mask data. 

\section{Benchmark Subtask Descriptions}
\label{app:subtasks}

We provide detailed descriptions of the subtask abbreviations used in Tab.~\ref{tab:dsrbench} and Tab.~\ref{tab:dynbench}.

\paragraph{DSR-Bench subtasks.}
DSR-Bench~\cite{dsrsuite} organizes its 13 subtasks along two axes: \emph{viewpoint mobility} (Absolute vs.\ Relative) and \emph{spatial attribute type}. ``Absolute'' (A.) denotes that the viewpoint is fixed at a specific timestamp, while ``Relative'' (R.) denotes that the viewpoint moves with the observing agent over time. The attribute types are:
\begin{itemize}[nosep,leftmargin=1.5em]
    \item \textbf{Dis} (Distance): How the distance between objects changes over time.
    \item \textbf{Dir} (Direction): The movement direction of a target object.
    \item \textbf{Ori} (Orientation): How the orientation of an object evolves.
    \item \textbf{Spd} (Speed): The speed change pattern of a target object.
    \item \textbf{SpdC} (Speed Comparison): Comparing the speeds of two objects.
    \item \textbf{DirP} (Direction Prediction): Predicting the future movement direction.
\end{itemize}
The 13th subtask, \textbf{N-Temp} (Non-Template Based), consists of free-form questions auto-generated by a language model to probe more general spatial-temporal understanding beyond fixed templates.

\paragraph{Dyn-Bench subtasks.}
Dyn-Bench~\cite{thinkdynamic} structures evaluation along three complementary semantic axes:
\begin{itemize}[nosep,leftmargin=1.5em]
    \item \textbf{Inter-Object}: How multiple dynamic objects interact with each other, including spatial relations, approach/separation, occlusion, and action descriptions.
    \item \textbf{Object-Scene}: How individual objects move within their environment, covering movement patterns, trajectories, and scene-level dynamics.
    \item \textbf{Camera-Object}: How camera motion affects the perceived geometry and temporal consistency of dynamic objects, including camera motion orientation, camera-object interaction, as well as the temporal visual changes.
\end{itemize}

\begin{figure*}[htbp]
    \centering
    \includegraphics[width=\textwidth]{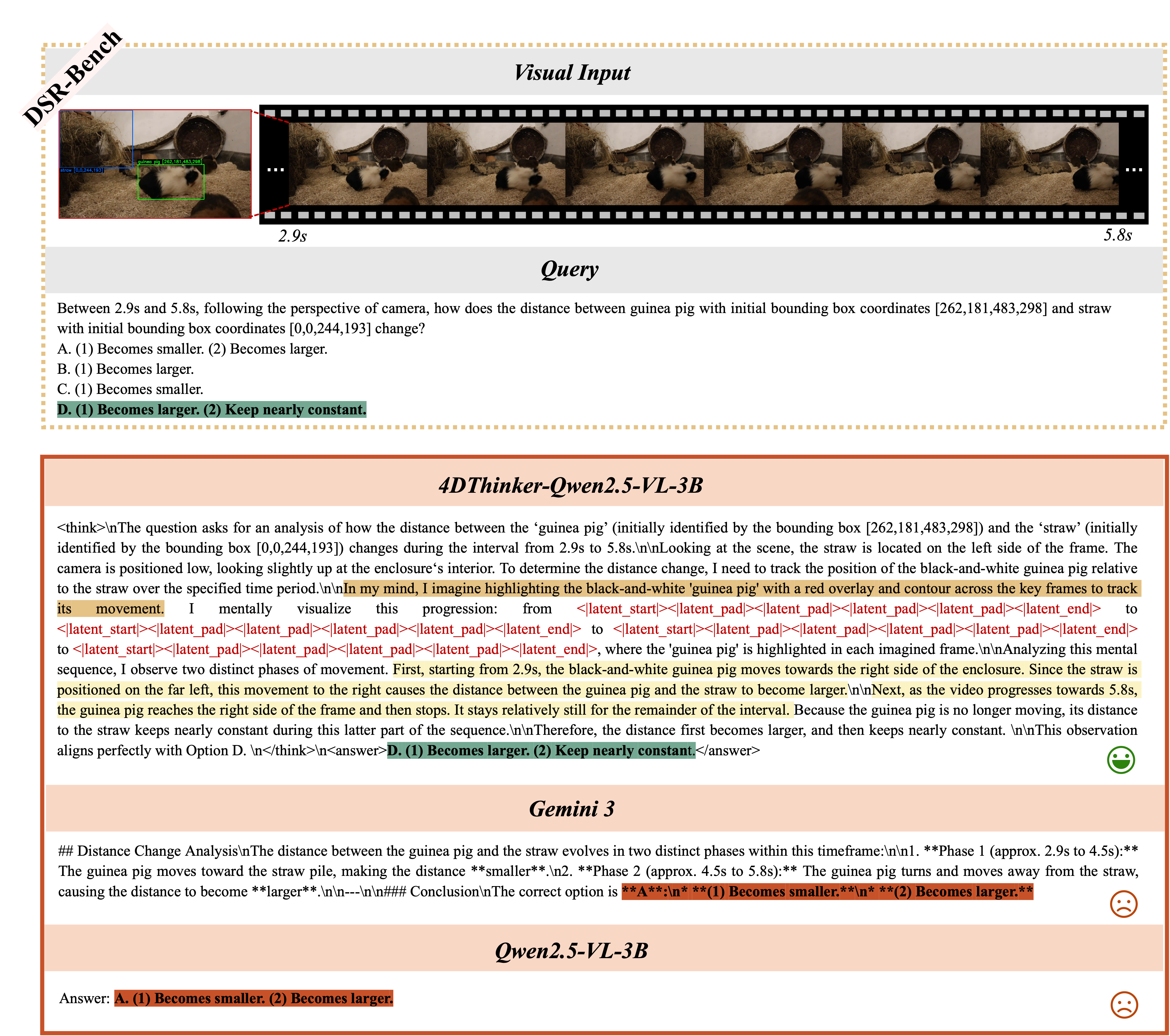}
    \caption{\textbf{Qualitative example on DSR-Bench (fine-grained).} \method correctly identifies a two-phase pattern (first becomes larger, then keeps constant) by mentally simulating the guinea pig's trajectory via latent 4D imagery. Both Gemini-3 and the base Qwen2.5-VL-3B fail.}
    \label{fig:showcase1}
\end{figure*}

\begin{figure*}[htbp]
    \centering
    \includegraphics[width=\textwidth]{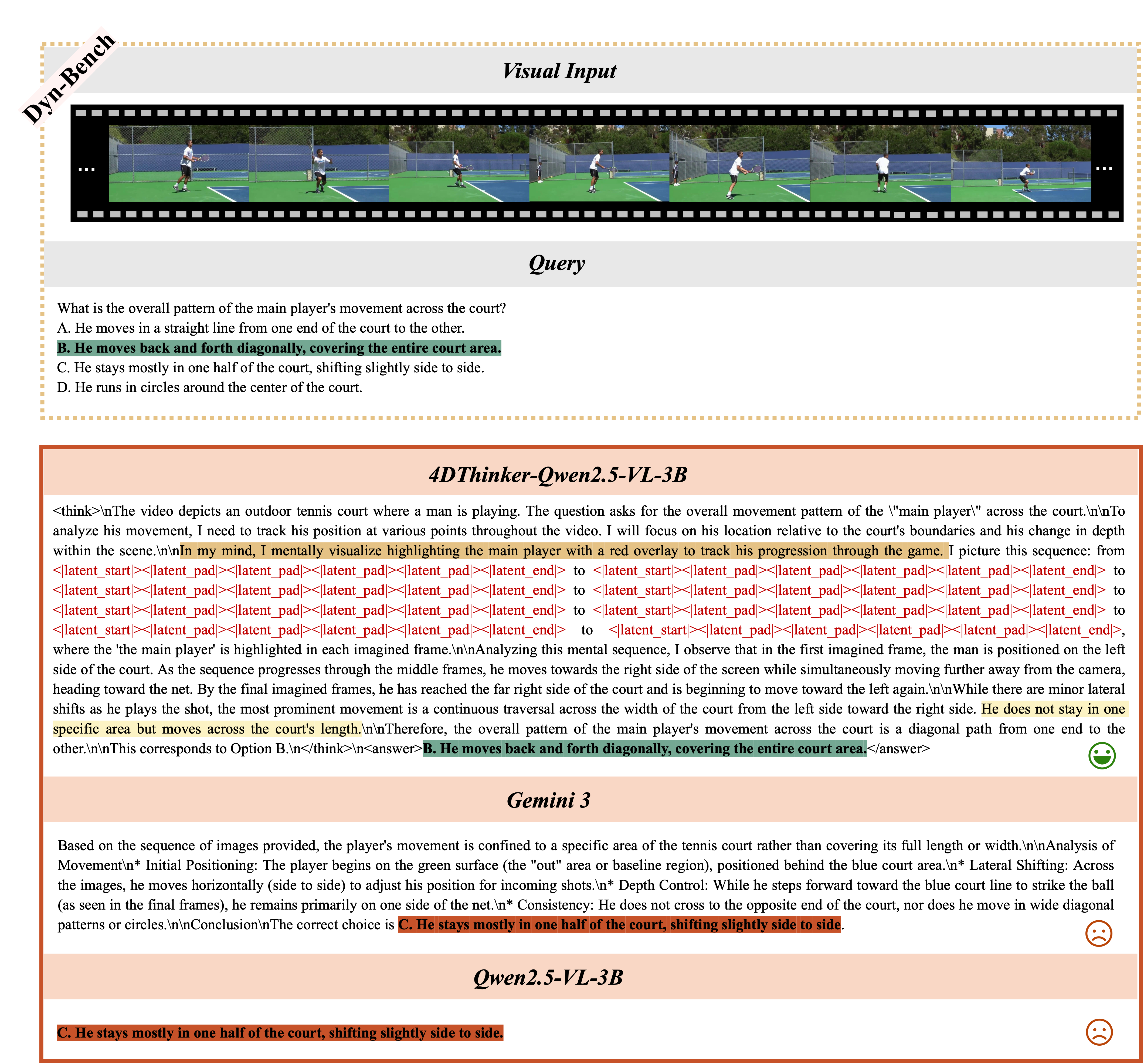}
    \caption{\textbf{Qualitative example on Dyn-Bench (holistic).} \method correctly identifies the player's diagonal movement pattern across the full court by mentally tracking his position through 4D latents, while both Gemini-3 and the base Qwen2.5-VL-3B incorrectly conclude that the player stays in one half of the court, relying on local frame-level heuristics.}
    \label{fig:showcase2}
\end{figure*}

\begin{figure*}[htbp]
    \centering
    \includegraphics[width=\textwidth]{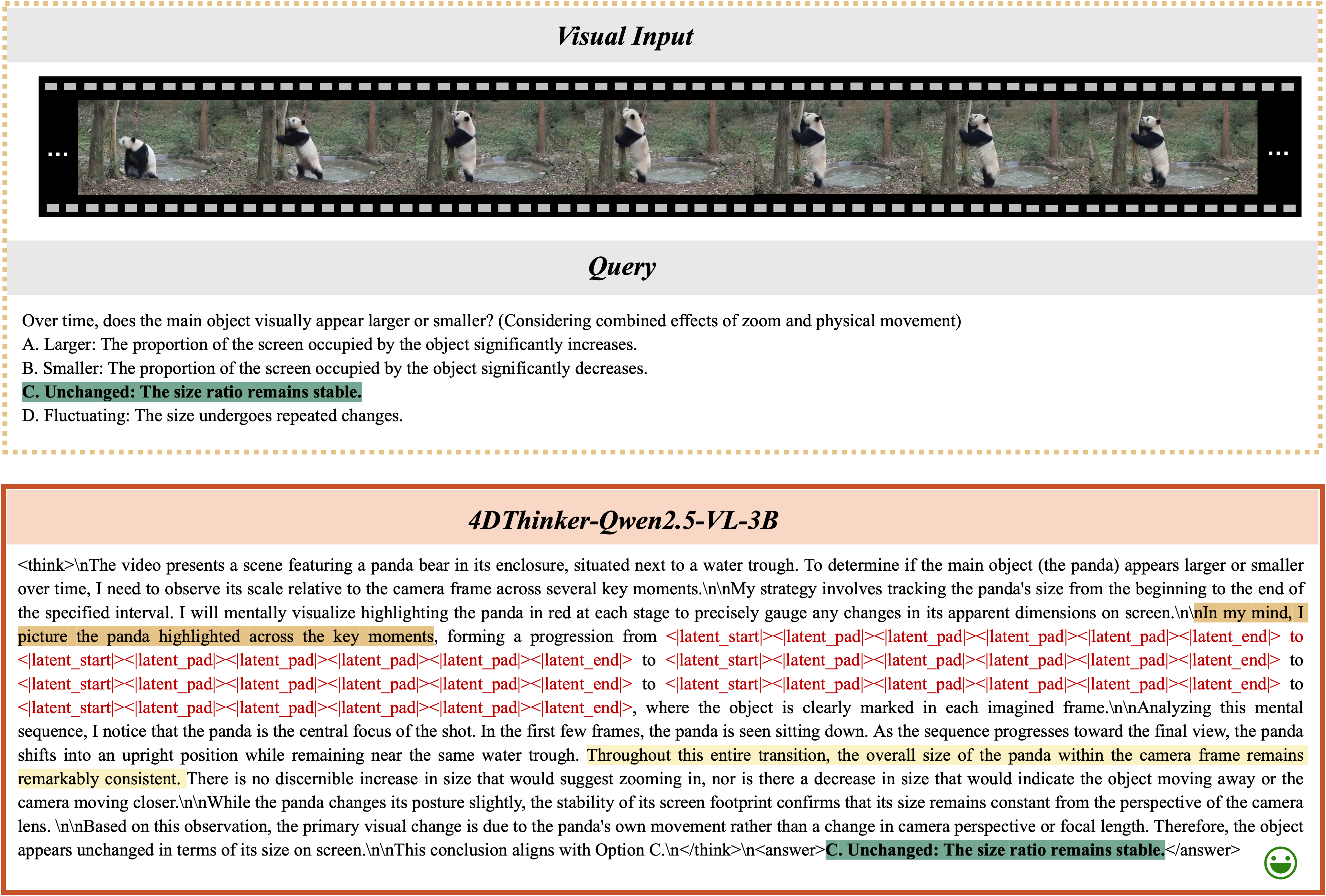}
    \caption{\textbf{Qualitative example.} \method tracks the panda's apparent size across frames through latent visual tokens and correctly determines that the size ratio remains stable, distinguishing the panda's posture change from actual camera zoom or physical depth movement.}
    \label{fig:showcase3}
\end{figure*}

\begin{figure*}[htbp]
    \centering
    \includegraphics[width=\textwidth]{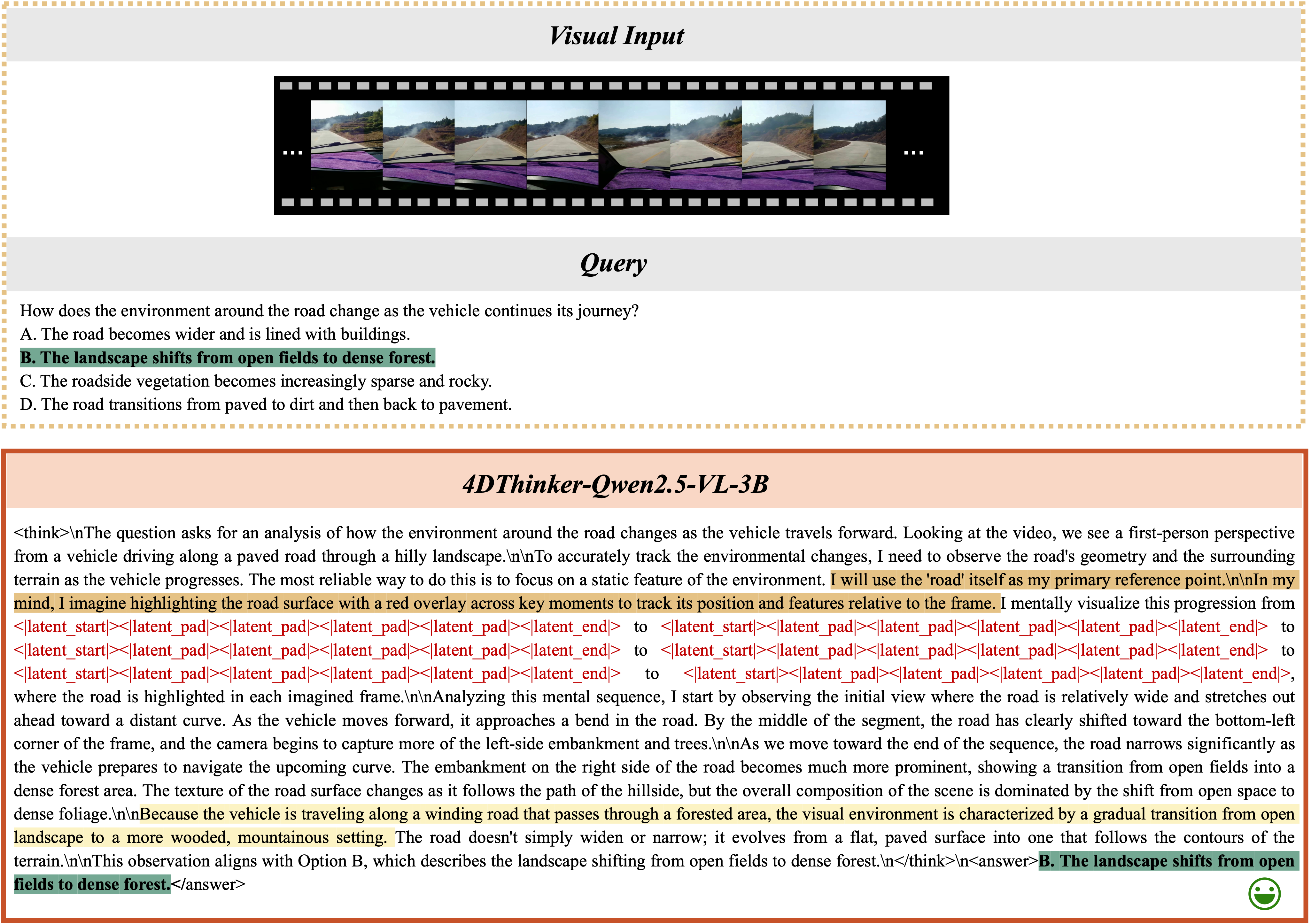}
    \caption{\textbf{Qualitative example.} Given a first-person driving video, \method tracks gradual environmental transitions via 4D latents and predict the open fields to dense forest.}
    \label{fig:showcase4}
\end{figure*}

\section{Automated CoT Validation}
\label{app:cot}

To ensure data quality, we apply a rule-based validator mentioned in Sec.~\ref{sec:data} that checks:
\begin{itemize}[nosep,leftmargin=1.5em]
    \item \textbf{Format completeness}: The response must contain properly paired \texttt{<think>...</think>} and \texttt{<answer>...</answer>} tags.
    \item \textbf{Placeholder count}: The number of placeholders must match the number of overlay images.
    \item \textbf{Answer validity}: The generated answer must exactly match one of the predefined options.
\end{itemize}
Samples failing any check are regenerated up to three times; persistent failures are discarded.

\section{Qualitative Examples}
\label{app:showcase}

We present qualitative examples in Figs.~\ref{fig:showcase1}--\ref{fig:showcase4} to illustrate how \method leverages its ``think with 4D'' reasoning process. In each example, the model first articulates the reasoning strategy with a sequence of latent visual tokens (displayed as \texttt{<|latent\_start|>...<|latent\_end|>}) representing its internal 4D mental imagery, and then derives the answer.

Fig.~\ref{fig:showcase1} demonstrates a DSR-Bench example requiring procedural distance tracking, where \method correctly decomposes the motion into two temporal phases via latent imagery. Fig.~\ref{fig:showcase2} shows a Dyn-Bench example where \method captures the global movement pattern (diagonal traversal) that confuses models (e.g., Gemini-3, Qwen2.5-VL-3B) relying on local frame-level heuristics. Fig.~\ref{fig:showcase3} illustrates \method's ability to \emph{disentangle object posture changes} from camera-induced size variations through continuous tracking in latent space. Fig.~\ref{fig:showcase4} shows that \method can anchor on static reference landmarks to \emph{capture gradual scene-level environmental transitions and predict the future scene} in egocentric driving videos.

\section{Implementation Details}
\label{app:implementation}

\paragraph{Base model.}
We build \method on a list of base VLM (e.g., Qwen2.5-VL~\cite{bai2025qwen2.5}, Qwen3-VL~\cite{qwen3technicalreport}, InternVL3.5~\cite{wang2025internvl3_5}, etc.). The visual encoder is kept \emph{frozen} throughout all training stages.

\paragraph{DIFT training.}
In terms of Qwen2.5-VL-3B, we train for 1 epochs with a batch size of 1. We use the AdamW optimizer with a learning rate of $1 \times 10^{-5}$ and latent size 4. The loss weights are set to $\lambda_{\text{ce}} = 0.1$ and $\lambda_{\text{sim}} = 1.0$. Input videos are uniformly sampled to 1 FPS. Training is conducted on 8 NVIDIA H200 141GB GPUs using DeepSpeed ZeRO-2.

Note that training configurations vary slightly across different foundation models. We utilized up to 64 NVIDIA H200 141GB GPUs to train a single model (e.g., InternVL3.5-38B).

\paragraph{4DRL training.}
We apply modified GRPO with the DSR-Train dataset. In terms of Qwen2.5-VL-3B, the group size is $G{=}8$. The reward weights are $\lambda_{\text{acc}} = 1.0$ and $\lambda_{\text{fmt}} = 0.2$. We use a learning rate of $1 \times 10^{-6}$ with the max completion length $= 8192$ and KL coefficient $\beta = 0.01$. Additionally, the batch size is 8 with 2 gradient accumulation steps. 

Configurations also vary slightly across models, with training requiring up to 64 NVIDIA H200 141GB GPUs per model. Note that specific RL hyperparameters exert a disproportionately large influence on the overall training process.

\paragraph{Mask overlay parameters.}
For generating mask overlays (Eq.~\eqref{eq:overlay}), we use opacity $\alpha = 0.6$ and highlight color $\mathbf{c} = [255, 0, 0]$ (red) for both static and dynamic objects.



\end{document}